\documentclass[conference]{IEEEtran}

\usepackage{layout,color,times,helvet,multicol,multirow,lscape,rotating,fancyhdr,balance}
\usepackage{algorithm}
\usepackage{algorithmic}
\usepackage{graphicx}
\usepackage{setspace}
\usepackage{color}
\usepackage{amsmath}
\usepackage{lineno}
\usepackage{colortbl}

\begin{document}

\title{Evolving Fuzzy Image Segmentation \\with Self-Configuration}

\author{
\IEEEauthorblockN{A. Othman$^1$, H.R. Tizhoosh$^2$, F. Khalvati$^3$}
\IEEEauthorblockA{$^1$ Dept. of Information Systems, Computers \& Informatics, 
 Suez Canal University, Egypt :: a.othman@ci.suez.edu.eg\\ $^2$ Centre for Pattern Analysis and Machine Intelligence, 
University of Waterloo, Canada :: tizhoosh@uwaterloo.ca\\ $^3$ Sunnybrook Health Sciences Centre, University of Toronto, Canada :: farzad.khalvati@sri.utoronto.ca}
}

\maketitle

\begin{abstract}
Current image segmentation techniques usually require that the user tune several parameters in order to obtain maximum segmentation accuracy, a computationally inefficient approach, especially when a large number of images must be processed sequentially in daily practice. The use of evolving fuzzy systems for designing a method that automatically adjusts parameters to segment medical images according to the quality expectation of expert users has been proposed recently (Evolving fuzzy image segmentation -- EFIS). However, EFIS suffers from a few limitations when used in practice mainly due to some fixed parameters. For instance, EFIS depends on auto-detection of the object of interest for feature calculation, a task that is highly application-dependent. This shortcoming limits the applicability of EFIS, which was proposed with the ultimate goal of offering a generic but adjustable segmentation scheme. In this paper, a new version of EFIS is proposed to overcome these limitations. The new EFIS, called self-configuring EFIS (SC-EFIS), uses available training data to self-estimate the parameters that are fixed in EFIS. As well, the proposed SC-EFIS relies on a feature selection process that does not require auto-detection of an ROI. The proposed SC-EFIS was evaluated using the same segmentation algorithms and the same dataset as for EFIS. The results show that SC-EFIS can provide the same results as EFIS but with a higher level of automation.
\end{abstract}
\maketitle

%-----------------
\section{introduction}
Evolving fuzzy image segmentation (short EFIS \cite{EFIS}) has been recently introduced to solve the parameter setting problem (e.g., fine-tuning) of different segmentation techniques. EFIS has been designed with emphasis on acquiring and integrating user feedback into the fine-tuning process. As a result, EFIS is suitable for all applications, such as medical image analysis, in which an experienced and knowledgeable user provides evaluative feedback of some sort with respect to the quality, i.e., accuracy, of the image segmentation.

Image segmentation is the grouping of pixels to form meaningful clusters of pixels that constitute objects (e.g., organs, tumours), a task with various applications in medical image analysis including measurement, detection, and diagnosis. Image segmentation can be roughly categorized into two main classes of algorithms; non-parametric-based (e.g., atlas-based segmentation) and parametric-based (e.g., thresholding, region growing) algorithms. The former is based on a model which usually does not require parameters whereas the latter is based on some parameters that must be adjusted in order to obtain reasonable segmentation results. Parameter-based segmentation algorithms always face the challenge of parameter adjustment; a parameter tuned for a particular set of images may perform poorly for a different image category.

On the other hand, in a clinical setting such as in a hospital, the final outcome of image segmentation algorithms usually need to be modified (i.e., manually edited) and approved by a an expert (e.g., radiologist, oncologist, pathologist). The clinical ramifications of not verifying the correctness of segments include missing a target (resulting in a less effective therapy) or increased toxicity if the target is over-segmented. The frequent expert intervention to correct the results, in fact, generates valuable feedback for a learning scheme to automatically adjust the segmentation parameters. 

EFIS is an images segmentation scheme that evolves fuzzy rules to tune the parameters of a given segmentation algorithm by incorporating the user feedback which is provided to the system as corrected or manually created segmentation results called \emph{gold standard images}. EFIS represents a new understanding of how image segmentation should be designed in the context of observer-oriented applications. Naturally, EFIS needs to be further improved and extended in order to exploit the full potential of its underlaying evolving mechanism in relation to the user feedback. The original design of EFIS as presented in~\cite{EFIS} requires pre-configurations of a few steps which should be set for a given image set and the segmentation algorithm to which EFIS is integrated. This limits the efficiency of EFIS; either the algorithm should be pre-configured for each dataset and/or segmentation algorithm or it is possible that a fixed pre-configuration will adversely affect its performance. In this paper, we present a new and extended version of EFIS which we call self-configuring EFIS (short SC-EFIS) that has a higher level of automation. The new extension of EFIS proposed in this paper will enhance EFIS through removing these limitations by introducing self-configuration into different stages of EFIS.

This paper is organized as follow: In section~\ref{SummaryEFIS}, a brief review of the EFIS (evolving fuzzy image segmentation) will be provided. In section \ref{criticEFIS}, we critically point to the shortcomings of EFIS. The section \ref{featureSELECT} reviews the literature on feature selection as this is the major improvement in SC-EFIS compared to EFIS. In section~\ref{SCEFISsection}, we present the proposed self-configuring EFIS (SC-EFIS). In section~\ref{expResults}, experiments are described and the results are presented and analyzed. Finally, section~\ref{CON} concludes the paper.

%*************************************************
%*************************************************
%*************************************************
\section{A Brief Review of EFIS}
\label{SummaryEFIS}

The concept of Evolving Fuzzy Image Segmentation, EFIS, was proposed recently \cite{EFIS}. The problem that EFIS attempts to address is parameter adjustment in image segmentation. The basic idea of EFIS is to adjust the parameters of segmentation to increase the accuracy by using user feedback in form of corrected segments. To do so, EFIS extracts features from a region inside the image  and assigns them to the best parameter exhaustively detected. Clustering or other methods are then used to generate fuzzy rules, which are then continuously updated when new images are processed. The simplified pseudo-code of EFIS is given in Algorithm \ref{EFISsimple}.

%----------------------- SIMPLIFIED EFIS--------------------------------------
 % ALGORITHM Train EFIS -------------------------
\begin{algorithm}[t]
\caption{EFIS \cite{EFIS}: Simplified Overview}
\begin{algorithmic}[h]
\label{EFISsimple}
\STATE \textbf{------------ Training: Stage 1 ------------}
\STATE Determine the parent algorithms and their parameters 
\STATE Read the training images and their gold standard images 
\STATE Via exhaustive/trial-and-error comparisons with gold standard images, determine the best segments and the best parameter(s) that generate the best segments
\STATE \textbf{------------ Training: Stage 2 ------------}
\STATE Read the available training images 
\STATE Determine regions of interest (ROIs) around each segment 
\STATE Save ROIs for  each image 
\STATE \textbf{------------ Training: Stage 3 ------------} \\
\STATE Set the number of seeds inside the segments, and the number of rules to be extracted
\FOR  { all images}
	\FOR { all seeds}
		\STATE Determine a new seed point inside the ROI
		\STATE Extract features from the seed point's neighbourhood
         	\STATE Save features and best parameters in matrix $M$
	\ENDFOR
\ENDFOR
\STATE Generate fuzzy rules from the rule matrix $M$ 
\STATE Save the rule matrix $M$ and the generated rules
\STATE \textbf{------------ Online: Evolving Phase ------------} \\
\STATE Load the fuzzy rules and the rule matrix $M$ 
\STATE Read a new image 
\STATE Detect ROI
\STATE Determine seed points inside ROI
\STATE Extract features from the seed point's neighbourhood 
\STATE Perform fuzzy inference to generate output(s): $\textrm{parameters} =$ FUZZY-INFERENCE(\textrm{RULES})
\STATE Apply the parameters to segment the image
\STATE Display the segment and wait for the user feedback (user generates a gold standard image by editing the segment)
\STATE \textbf{----------- *Rule Evolution - Invisible to User* -----------}
\STATE Determine the best output(s) (via comparison of segments with the gold standard image)
\IF {(Pruning) the features/parameters not seen yet} 
	\STATE Add new rows to the rule matrix
	\STATE Generate fuzzy rules from the rule matrix $M$
	\STATE Save the rule matrix $M$ and the generated rules
\ENDIF
\end{algorithmic}
 \end{algorithm}
%-------------------------------------------------------------------------------------- 

EFIS needs to be trained for specific algorithms and image categories~\cite{EFIS}. In other words, in order to employ EFIS, the following components must be pre-designated:
\begin{itemize}
\item Parent algorithm: any segmentation algorithm with at least one parameter that affects its  accuracy (e.g., global thresholding, statistical region merging)
\item Parameter(s) to be adjusted (e.g., thresholds, scales)
\item Images and corresponding gold standard images 
\item Procedure to find optimal parameters (e.g., brute force or trial-and-error via comparison with the gold standard images)
\end{itemize}
Once the above-mentioned components are available/defined, the following steps need to be specified:
\begin{itemize}
\item ROI-detection algorithm: An algorithm that detects the region of interest (ROI) around the subject to be segmented by EFIS.
\item Procedure for feature extraction around available seed points: Methods like SIFT are used to generate seed points. But a certain number of expressive features should be calculated in the vicinity of each seed point to be fed to fuzzy inference system.
\item Rule pruning: Upon processing a new image, a new rule can be learned only if the features and corresponding output parameters had not been observed previously. In other words, by looking at the difference between an input  (features plus outputs) with all rules in the database, the information of a new image is added only if not captured by existing rules.
\item Label fusion: When EFIS is used with multiple algorithms at once, the segmentation results are fused using a fusion method namely STAPLE algorithm~\cite{Warfield2004}.
\end{itemize}

EFIS includes two main phases namely training and testing. In training phase, images with their gold standard results are fed to the algorithm where features are extracted from each image. The parent algorithm, e.g., thresholding, is applied to each image and the results are compared to the gold standard image. The algorithm's parameters are continuously changed until the best possible result is achieved. The best parameter which yields the best result (i.e., the highest agreement with the gold standard image) along with the image feature extracted in the previous stage are stored. Once all training images are processed, the fuzzy rules are generated from the stored data using a clustering algorithm.

In testing phase, new images are first processed to extract features. Next, the image features are fed to the fuzzy inference system to approximate the parameters. The parent algorithm is then applied to the input image using the estimated parameter. EFIS can address both single-parametric and multi-parametric problems. EFIS was applied to three different thresholding algorithms  and significant improvements in terms of segmentation accuracy were achieved \cite{EFIS}.

%*************************************************
%*************************************************
%*************************************************
\section{Critical Analysis of EFIS}
\label{criticEFIS}
Although EFIS has demonstrated to improve the segmentation results~\cite{EFIS}, some of its underlying steps may limit its applicability mainly because these steps have been designed in an ad-hoc fashion and tailored to the specific test images and algorithms namely breast ultrasound and thresholding. In this section, we examine the limitations of EFIS and lay out how they should be addressed via self-configuration.

EFIS calculates the features inside a rectangle that constitutes the region of interest, ROI. Within this region, $n$ feature are calculated using scale-invariant feature transform (SIFT)~\cite{lowe1999object,lowe2004distinctive}. In designing the ROI-detection algorithm, it is assumed that the ROI will be dark based on the characteristics of test images used (breast lesions in ultrasound are hypoechoic, meaning they are darker than surrounding tissue). This means that EFIS needs a detection algorithm for any new image category (application) to correctly recognize the region of the image containing the object of interest. In addition, similar to any other detection algorithm, if it fails, then EFIS will not be able to perform. We will remove this dependency by redesigning the feature extraction stage.

In order to calculate features within the ROI, EFIS uses a fixed number of landmarks, called seed points, which are delivered by SIFT.  These $n$ fixed key points, with $n=10$, is set for all images regardless of their content. Of course, an arbitrary number of features may not be able to characterize all types of images. We will eliminate this limitation of EFIS by automatically setting the number of seed points for different image categories.

EFIS constructs a fixed sized window of $40 \times 40$ pixels around each landmark (seed point) to calculate the features. A self-configuring EFIS  has to automatically set the window size during a pre-processing stage in order to optimally define the feature neighbourhood.

EFIS uses a fixed number of manually selected features, namely 18 features which proved to perform well on the breast ultrasound images. It is intuitively clear that this may not be a flexible approach to capture the image content. Any set of images with some common characteristics may need a different set of features for the evolving fuzzy systems to effectively estimate the parameters of the segmentation.

In the proposed extension of EFIS algorithm, we will address these shortcomings by introducing a pre-processing (self-configuration) stage where the settings are undertaken automatically. As apparent from the list above, feature selection seems to be the core of EFIS lack of automation. In following section, therefore, we will briefly review feature selection methods.

%*************************************************
%*************************************************
%*************************************************
\section{Feature Selection}
\label{featureSELECT}

Providing relevant features to a learning system will increase its ability to generalize and hence elevate its performance. Feature selection is the process of selecting the most relevant features out of a larger group of features so that either redundant or irrelevant features are removed. Redundant features add no new information to the system, and irrelevant features may confuse the system and decrease its ability to learn efficiently. Feature selection may be conducted according to one of four schemes \cite{molina2002feature}:

\begin{itemize}
\item \textbf{Filter feature selection} methods work directly on the available data and select features based on the data properties. They are independent of any learning methods \cite{sanchez2007filter,lal2006embedded,Arvacheh2005}.

\item \textbf{Wrapper feature selection} methods may evaluate features but without consideration of the structure of the classifier~\cite{lal2006embedded}.

\item \textbf{Embedded feature selection} treats the learning and feature selection aspects as one process.

\item \textbf{Hybrid systems} may combine wrapper and filter approaches~\cite{Cad2013}.
\end{itemize}

Feature selection may also be categorized into three main branches: supervised, semi-supervised, and unsupervised.

\subsection{ Supervised Feature Selection}
In supervised feature extraction, the selection of a set of features from a larger number of features is based on one of three characteristics~\cite{molina2002feature}: 1) features of a size that optimize an evaluation measure, 2) features satisfying a condition in the evaluation measure, and 3) features that best match a size and evaluation measure. Supervised feature selection methods deal primarily with the classification problems, in which the class labels are known in advance~\cite{saeys2007review}. Numerous studies have investigated supervised feature selection using the measures of the information theoretic~\cite{martinez2010supervised} and Hilbert-Schmidt independence criterion~\cite{song2007supervised}.

\subsection{Semi-Supervised Feature Selection}
The concept of semi-supervised feature selection has emerged recently as a means of addressing situations in which insufficient labels are available to cover the entire training data \cite{zhao2007semi} or in which a substantial portion of the data are unlabelled. Traditional supervised feature selection techniques are generally ineffective under such circumstance. Semi-supervised feature selection is therefore employed for the selection of features when not enough labels are available. A semi-supervised feature selection constraint score that takes into account the unlabelled data has been proposed in \cite{kalakech2011constraint}. The literature also contains proposals for numerous semi-supervised techniques based on spectral analysis \cite{zhao2007semi}, a Bayesian network \cite{cai2011bassum}, a combination of a traditional technique with feature importance measure \cite{bellal2012semi}, or the use of a Laplacian score \cite{doquire2013graph}. Although semi-supervised selection does not require a complete set of class labels, it  does need some. 

\subsection{ Unsupervised Feature Selection}

Unsupervised feature selection is the process of selecting the most relevant non-redundant features from a larger number of features without the use of class labels. Mitra et al. \cite{mitra2002unsupervised} proposed an unsupervised feature selection algorithm based on feature similarity. They used a maximum information compression index to measure the similarities between features so that similar features could be discarded. He et al. \cite{he2006laplacian} proposed an unsupervised feature selection technique that relies on the Laplacian score to indicate the significance of the features.  Zhao et al. \cite{zhao2007spectral} used spectral graph theory to develop a new algorithm that unifies both supervised and unsupervised feature selection in one algorithm. They applied the spectrum of the graph that contains the information about the structure of the graph in order to measure the relevance of the features. Cai et al. \cite{cai2010unsupervised} proposed a new unsupervised feature selection algorithm called Multi-Cluster Feature Selection, in which the features selected are those that maintain the multi-cluster structure of the data. Farahat et al. \cite{farahat2012efficient} present a novel unsupervised greedy feature selection algorithm consisting of two parts: a recursive technique for calculating the reconstruction error of the matrix of features selected, and a greedy algorithm for feature selection. The method was tested on six different benchmark data sets, and the results show an improvement over state-of-the-art unsupervised feature selection techniques.

\subsection{Features for SC-EFIS}
In order to eliminate the major shortcomings of EFIS with respect to inflexible and static feature selection, and in order to not assume availability of class labels, we chose unsupervised feature selection, specially the previously mentioned five popular unsupervised feature selection algorithms to characterize images for training the evolving fuzzy system. These five methods, along with an additional correlation-based method, were combined to produce an ensemble of final relevant features that could be used for training.

In the remaining of the paper, the output matrices of these techniques are denoted as follows:
\begin{itemize}
\item Mitra et al. \cite{mitra2002unsupervised}- $F_F$ (feature similarity).
\item He et al. \cite{he2006laplacian}- $F_L$ (Laplacian score).
\item Zhao et al. \cite{zhao2007spectral}- $F_P$ (spectral graph).
\item Cai et al. \cite{cai2010unsupervised}- $F_M$ (multi-cluster).
\item Farahat et al. \cite{farahat2012efficient}- $F_G$ (greedy algorithm).
\item $F_C$ (correlation method).
\end{itemize}
%-----------------------

%*************************************************
%*************************************************
%*************************************************
\section{Self-Configuring EFIS (SC-EFIS)}
\label{SCEFISsection}
This section introduces a new version of EFIS, namely a self-configuring evolving fuzzy image segmentation (SC-EFIS) which represents a higher level of automation compared to the original EFIS scheme. The proposed SC-EFIS scheme consists of three phases;  self-configuration phase, training phase, and online or evolving phase. In the following, each of these phases are described in detail.

\subsection{Self-Configuring Phase}
\label{SCEFISPre}

In the self-configuring phase (Algorithm \ref{SCEFIS_Pre}), all available images are processed in order to determine two crucial factors: 1) the size of the feature area around each seed point, and 2) the final features to be used for the current image category.

\begin{figure}[htb]
\includegraphics[width=1\columnwidth]{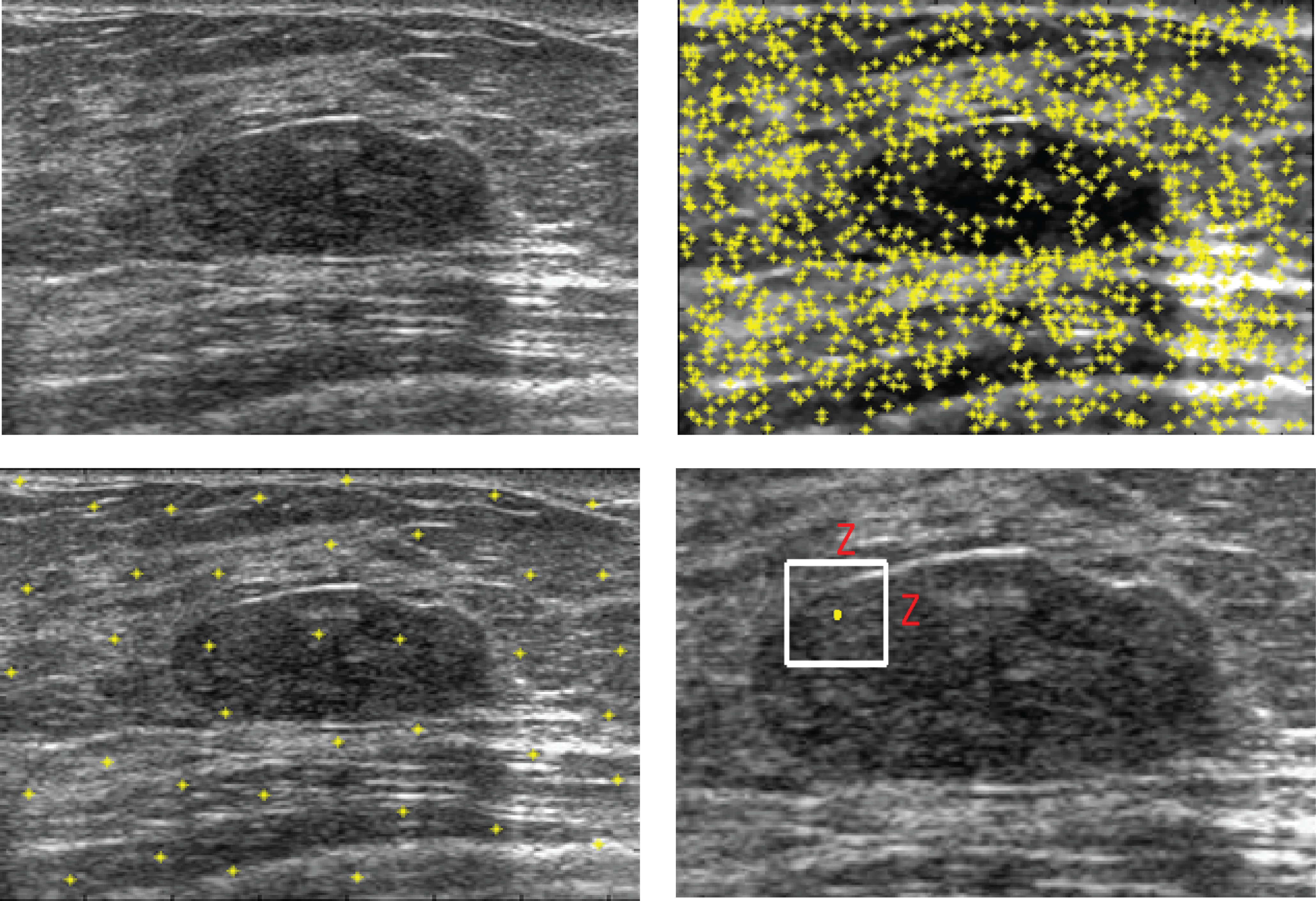}
\caption{Feature extraction process (from top left to bottom right): original image, seed points detected by SIFT, selected seed pints via sorting the descriptor, calculating features around each selected seed point.}
\label{Extract}
\end{figure}

The $Z \times Z$ rectangle around each SIFT point to be used for feature calculation is determined based on different sizes of all available images (algorithm \ref{SCEFIS_Pre}). Following this step, the set of features that should be used for the available images is selected from a large number of features which are calculated for each image from the vicinity of the SIFT points located in the entire image (since there is no longer an ROI) (Fig. \ref{Extract}). This process starts with the determination of the number of SIFT points $N_F$ that should be used in the current image (algorithm \ref{SCEFIS_Pre}). This step is identical to the procedure used in the EFIS training phase, as previously explained in section~\ref{SummaryEFIS}, with three exceptions: the SIFT points are detected across the entire image (as opposed to selecting SIFT points inside an ROI as a subset of the image), the final number $N_F$ of SIFT seed points is not fixed, and the points returned are separated from each other by $Z$ in each direction. For all $N_F$ seed points, features are extracted from a rectangle $R_C$ around each point, based on the discrete cosine transform ($D_C$) of $R_C$, the gradient magnitude ($G_M$) of $R_C$, the approximation coefficient matrix $A_C$ of $R_C$ (computed using the wavelet decomposition of $R_C$), and the SIFT descriptors $D_S$. The following set of features is extracted (Algorithm \ref{SCEFIS_Pre}):

\begin{enumerate}
\item The mean, median, standard deviation, co-variance, mode, range, minimum, and maximum of $R_C$, $D_{C_{R_C}}$, and $A_{C_{R_C}}$, and $G_{M_{R_C}}$ (32 features)

\item The mean, median, standard deviation, co-variance, range, minimum, maximum, and zero population of $D_S$ (eight features) with the minimum of $D_S$ changed to be the minimum number after zero

\item The contrast, correlation, energy, and homogeneity of the gray level co-occurrence matrices (computed in four directions $0\,^{\circ}$, $45\,^{\circ}$, $90\,^{\circ}$, and $135\,^{\circ}$) of $R_C$, $D_{C_{R_C}}$, and $A_{C_{R_C}}$, and $G_{M_{R_C}}$ (64 features)

\item The contrast, correlation, energy, and homogeneity of the gray level co-occurrence matrices (computed in only one directions of $0\,^{\circ}$) of $D_S$ (four features)

\item A feature matrix $F_1$ of size $N_F \times N_T$ generated for $I$ (in this case $N_T=108$)
\end{enumerate}

% ALGORITHM Train EFIS -------------------------
\begin{algorithm}[htb]
\caption{\textcolor{black}{Self-Configuration Phase}}
\begin{algorithmic}[1]
\label{SCEFIS_Pre}
\STATE Set the variables and initialize all matrices 
\STATE Read the available images $I_1, I_2, \cdots , I_{N_I}$.
\STATE Read the size of the images, namely all rows $R_1, R_2, \cdots , R_{N_I}$, and all columns $C_1, C_2, \cdots , C_{N_I}$.
\STATE Determine the size of the rectangle\\ Z = $0.1\times \textrm{max}(\textrm{median}_i(R_i),\textrm{median}_i(C_i))$.
\STATE Create the initial matrix  $F_1$ and the final matrix $F^*$.
\FOR{each image}
\STATE Determine $N_F$, the number of SIFT points, that should be used for image $I_i$.
\FOR{each SIFT point}
\STATE Extract features $f_1,f_2,\cdots,f_{N_T}$  from the $Z \times Z$ rectangle around each SIFT point.
\STATE Append the features as a new row to the initial matrix $F_1$, which becomes of size $N_F \times N_T$.
  \ENDFOR
\STATE Calculate $S_T$ different statistics from $F_1$ and assigned in $F_2$.
\STATE Append $F_2$ of the current image of size $S_T\times N_T$ to the feature matrix $F_3$ (the feature matrix $F_3$ becomes of size $L \times N_T$, $ L= S_T*N_I$)
\ENDFOR
\STATE Remove very similar features from $F_3$ (e.g., at least 99\% correlated). $F_4$ is a reduced matrix of $F_3$ of size $L \times N_{T_1}$, $N_{T_1} \leq N_T$.
\STATE Determine the number of features by discarding similar ones from $F_4$ (e.g., at least 90\% correlated). $F_C$ is a feature matrix generated from $F_4$ of size $L \times N_{T_2}$, $N_{T_2} \leq N_{T_1}$.
\STATE Use $k$ different unsupervised feature selection methods to generate $k$ different feature matrices in addition to $F_C$: $F_P$, $F_M$, $F_F$, $F_G$, and $F_L$. All of these matrices are of size $L \times N_{T_2}$.
\STATE Select any features found in at least half of the matrices to form $F_5$ of size $L \times N_{T_3}$, $N_{T_3} \leq N_{T_2}$.
\STATE Generate a final feature matrix $F^*$ from $F_5$ by removing similar features (e.g., at least 90\% correlated). $F^*$ is of size $L \times N_L$, $N_L \leq N_{T_3}$.
 \end{algorithmic}
 \end{algorithm}
 
The next step is to calculate $S_T$ different statistical measures from $F_1$ (e.g.,  $S_T=8$: mean, median, mode, standard deviation, co-variance, range, minimum, and maximum). The resulting matrix $F_2$ (size $S_T \times N_T$) is returned, in which each row represents a statistical measure (Algorithm \ref{SCEFIS_Pre}, \emph{CSF}). $F_2$ is then appended to the feature matrix $F_3$ (Algorithm \ref{SCEFIS_Pre}). After all images are processed, the feature matrix $F_3$ is formed from the features of all images, with each image being represented by $S_T$ rows.

In the last step, the final set of features that should be used in the current image category are selected from $F_3$. This process starts with the removal of very similar features in $F_3$ based on the calculation of the correlations between all features. Hence, if two features are highly correlated, e.g. with a correlation coefficient of at least 99\%, then one is kept and the other is discarded. The output of this process is a matrix $F_4$  (Algorithm \ref{SCEFIS_Pre}).

%----------------2nd
 For any unsupervised feature selection technique, the number of features $N_{T_2}$ that should be returned must be established in advance. A correlation with a threshold of 90\% is used in order to determine the number of features that should be returned from $F_4$ (Algorithm \ref{SCEFIS_Pre}). Following this process, $F_C$ is the resulting feature matrix. In addition to $F_C$, five different unsupervised feature selection methods are also used for feature selection. The matrix $F_4$ and the variable $N_{T_2}$ are passed to the methods, and each method returns a different matrix with its selected features. The resulting matrices are $F_G$ \cite{farahat2012efficient}, $F_L$ \cite{he2006laplacian}, $F_F$ \cite{mitra2002unsupervised}, $F_P$ \cite{zhao2007spectral}, and $F_M$ \cite{cai2010unsupervised} (Algorithm \ref{SCEFIS_Pre}). For all features in the six matrices, any feature extracted by at least three of the six methods are selected and appended to a matrix $F_5$ (Algorithm \ref{SCEFIS_Pre}). The final matrix $F^*$ is generated based on the discarding of features from $F_5$ that are at least 90\% correlated (Algorithm \ref{SCEFIS_Pre}).
 
\subsection{Offline Phase}
In the offline phase, the best parameters for segmenting each image are calculated through an exhaustive search and then stored in matrix $T$ (Algorithm \ref{SCEFIS_TRAIN}, \emph{BSP}). The process is performed as explained in \cite{EFIS}.

\subsection{Training Phase}
\label{SCEFISTrain}
In this phase, the features selected for the training images are used for the training of the fuzzy system. A set of images are randomly selected for training (Algorithm \ref{SCEFIS_TRAIN}). A matrix $M$ is created and filled with the rows from $F^*$ that belong to the training images (Algorithm \ref{SCEFIS_TRAIN}). A matrix $O$ is created and filled with the rows from $T$ that belong to the training images (Algorithm \ref{SCEFIS_TRAIN}). A pruning step is performed starting from the second training image in order to ensure that $M$ and $O$ do not contain similar rows (Algorithm \ref{SCEFIS_TRAIN}). The pruned matrices $M$ and $O$ are used for the generation of the initial fuzzy rules (Algorithm \ref{SCEFIS_TRAIN}). The initial fuzzy system is built through the creation of a set of rules using the Takagi-Sugeno approach to describe the in- and output matrices. Based on $N_L$ different features from the input and one optimal parameter as the output, a set of rules is generated whereby the features are in the antecedent part and the optimal parameters are in the consequent part of the rules.

% ALGORITHM Train EFIS -------------------------
\begin{algorithm}[htbp]
\caption{Offline and Training Phases}
\begin{algorithmic}[1]
\label{SCEFIS_TRAIN}
\STATE \textbf{------------ Offline phase ------------}
\STATE Determine the parent algorithm(s) and their parameters $p_1,p_2,\cdots,p_k$.
\STATE Read the gold standard images $G_1, G_2, \cdots , G_n$.
\STATE Via exhaustive search or trial-and-error comparisons with gold standard images, determine the best segments $S_1, S_2, \cdots , S_n$ and the best parameters $p_1^*,p_2^*,\cdots,p_k^*$ that generate the best segments and store them in matrix $T$.
\STATE \textbf{------------ Training phase ------------}
\STATE Determine the available training images $I_1, I_2, \cdots , I_{N_R}$.
\STATE \textcolor{black}{Create two empty matrices $M$ for input and $O$ for output.}
\FOR  {all $N_R$ images}

		\STATE Fill matrix $F_T$ with rows from matrix $F^*$ that belong to the training image $I_i$
		($F_T = F^* (I_i) $).
		\STATE Fill matrix $T_R$ with rows from matrix $T$ that belong to the training image $I_i$\\
		($T_R = T(I_i) $).
		\IF {i=1}
		\STATE Append $F_R$ to $M$, and $T_R$ to $O$.
			\ELSE		
	 \STATE Pruning step: Discard rows from $F_R$ and $T_R$ that are similar to rows in $M$ and $O$, respectively.
	 \STATE Append the updated matrices $F_R$ and $T_R$ to $M$ and $O$ respectively.
		\ENDIF
	\ENDFOR

\STATE Generate fuzzy rules $R_{F_1},R_{F_2},\cdots$ from the input matrix $M$ and the output matrix $O$ (e.g., using clustering).

\end{algorithmic}
 \end{algorithm}
 
%----3rd Pahse-------------------------------------------------------------------------------
\subsection{Online and Evolving Phase}
\label{SCEFISTEst}

The evolving process is performed in order to increase the capabilities of the proposed system. For each test image, a matrix $F_S$ is filled with the rows from $F^*$ that belong to the test image (Algorithm \ref{SCEFIS_Use}). Fuzzy inference using $F_S$ is applied, and a parameter vector $T_O$ is returned (size $1 \times 8$) and the final output parameter $T^*$ is calculated (Algorithm \ref{SCEFIS_Use}). The resulting parameter is used for the segmentation of the image (Algorithm \ref{SCEFIS_Use}), and the resulting segment is stored and then displayed to the user for review and eventual correction (Algorithm \ref{SCEFIS_Use}). The best parameter for the current image is then calculated based on the user-corrected segment and is stored in $T_B$ (Algorithm \ref{SCEFIS_Use}). A pruning procedure is performed on $F_S$ and $T_B$ as described in \cite{EFIS}, with the exception that the Euclidean distance thresholds are, in contrast to EFIS, different for different techniques.  After pruning, revised versions of $F_S$ and $T_B$ are appended to $M$ and $O$ (Algorithm \ref{SCEFIS_Use}). In the final step, the current fuzzy inference system, i.e., its rule base, is regenerated using the updated matrices $M$ and $O$ (Algorithm \ref{SCEFIS_Use}), and the process is repeated as long as new images are available.

% ALGORITHM Use EFIS ------------------------
\begin{algorithm}[ht]
\caption{Online/Evolving Phase }
\begin{algorithmic}[1]
\label{SCEFIS_Use}
\STATE Load the fuzzy rules $R_{F_i}$ and the matrices $M$, $O$, and $F^*$.
\STATE Load the test images $I_1, I_2, \cdots , I_{N_E}$.
\FOR  { all $N_E$ images}
		\STATE Fill matrix $F_S$ with the rows from matrix $F^*$ that belong to the test image $I_i$ ($F_S = F^* (I_i) $).
\STATE Perform fuzzy inference to generate output:\\ $T_O =$ FUZZY-INFERENCE($R_{F_1},R_{F_2},\cdots$).
	
\STATE Generate a single output $T^*$ from $T_O$ using the mean of ${T_O}$ ($\mu_{T_O}$), the median of ${T_O}$ ($M_{T_O}$), the fuzzy membership ($m_{T_O}$) of the standard deviation of $T_O$ ($\sigma_{T_O}$) using a Z-shaped function ($zmf$) \\ $m_{T_O} =  zmf(\sigma_{T_O},[(\mu_{T_O}*0.10) \ \   (\mu_{T_O}*0.20)])$, and \\ $T^* = m_{T_O} * \mu_{T_O} + (1- m_{T_O}) * M_{T_O}$.
\STATE Apply the parameters to segment $I_i$.
\STATE Display segment $S$ and wait for user feedback (user generates a gold standard image $G$ by editing $S$)
\STATE \textbf{--------- *Rule Evolution - Invisible to User* ---------}
\STATE Determine the best output vector $p_1^*,p_2^*,\cdots,p_k^*$ (via comparison of $S$ with $G$) and store it in $T_B$.
 \STATE Pruning -- Discard rows from $F_S$ and $T_B$ that are similar to rows in $M$ and $O$, respectively.
	 \STATE Append the matrices $F_S$ and $T_B$ to $M$ and $O$, respectively.
	\STATE Generate fuzzy rules $R_{F_i}$ from the updated matrices $M$ and $O$ (e.g., using clustering).
	\ENDFOR
\end{algorithmic}
 \end{algorithm}

%*************************************************
%*************************************************
%*************************************************
\section{Experiments and Results}
\label{expResults}
%-- Experiments
This section describes the experiments conducted in order to test the proposed self-configuring EFIS (SC-EFIS). To build the initial fuzzy system, for each training set, a set of randomly selected images from the data set were used for the extraction of the features along with the optimum parameters as output. This initial fuzzy system was then used to test the proposed method using the remaining images. The initial fuzzy system evolves as long as new (unseen) images are fed into the system and as long as the segmentation results produced by the algorithms are corrected by an expert user in order to generate optimal parameter values. This process drives the evolution of the fuzzy rules for segmentation. During the experimentation, the training-testing cycle was repeated 10 times. The results of ten different trials for each segmentation technique and for each parent algorithm are presented in order to validate the performance of SC-EFIS. The number of rules was monitored during the evolution process in order to acquire empirical knowledge about the convergence of the evolving process.

The experimental results using an image dataset for three different segmentation techniques (region growing, global thresholding, and statistical region merging) are presented. All experiments were performed using Matlab 64-bit. 

\subsection{Image Data}
\label{Images}
\begin{figure*}[htb]
\center
\includegraphics[width=0.6in,height=0.6in]{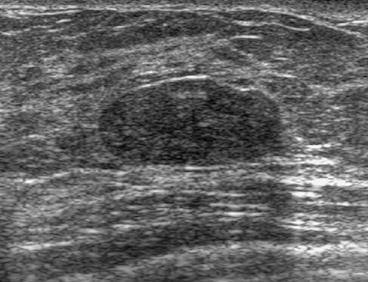}
\includegraphics[width=0.6in,height=0.6in]{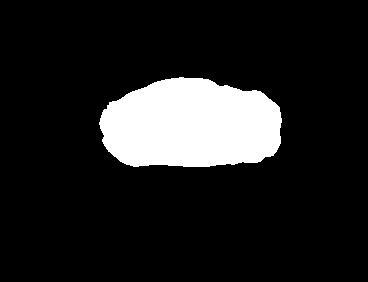}
\includegraphics[width=0.6in,height=0.6in]{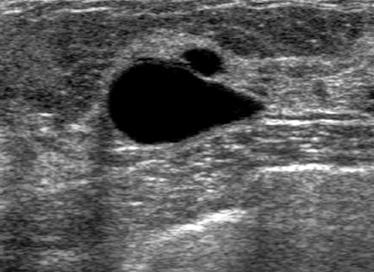}
\includegraphics[width=0.6in,height=0.6in]{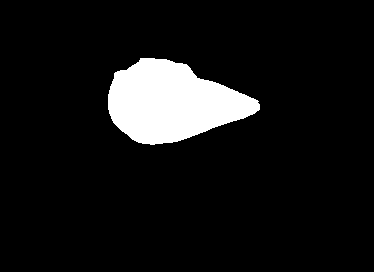}
\includegraphics[width=0.6in,height=0.6in]{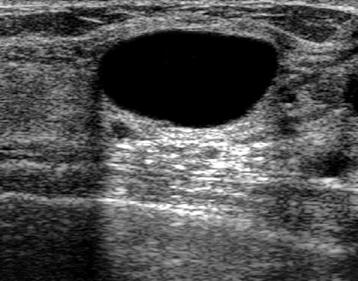}
\includegraphics[width=0.6in,height=0.6in]{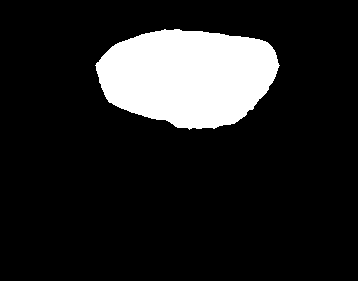}
\includegraphics[width=0.6in,height=0.6in]{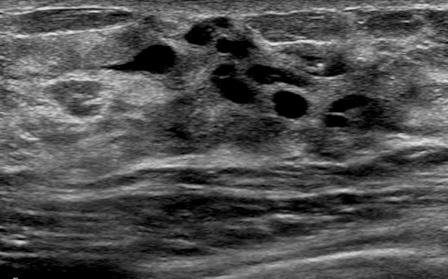}
\includegraphics[width=0.6in,height=0.6in]{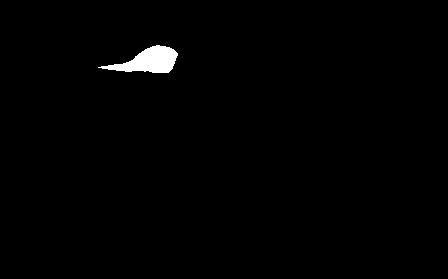}
\includegraphics[width=0.6in,height=0.6in]{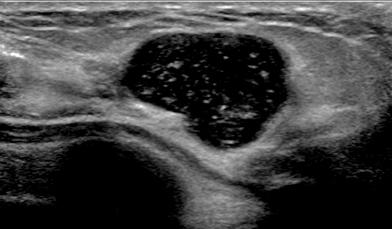}
\includegraphics[width=0.6in,height=0.6in]{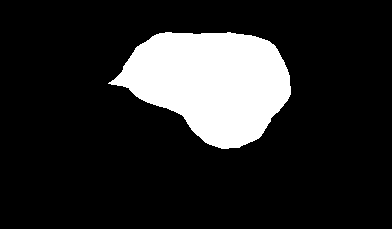} \\ \vspace{0.05in}
\includegraphics[width=0.6in,height=0.6in]{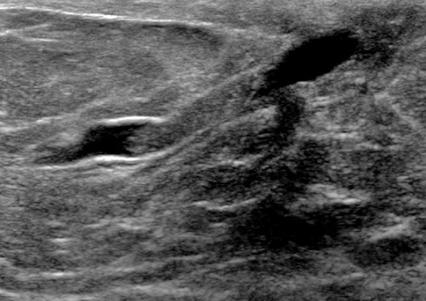}
\includegraphics[width=0.6in,height=0.6in]{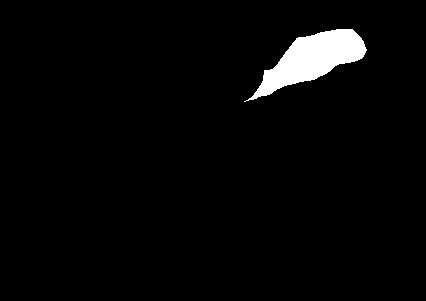}
\includegraphics[width=0.6in,height=0.6in]{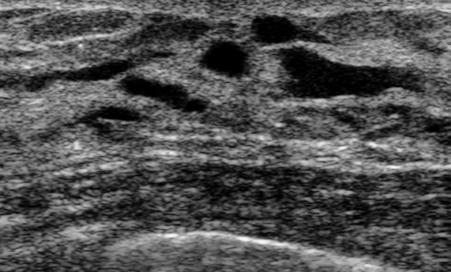}
\includegraphics[width=0.6in,height=0.6in]{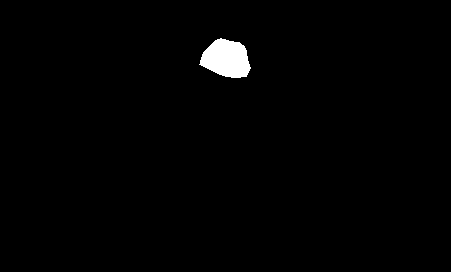}
\includegraphics[width=0.6in,height=0.6in]{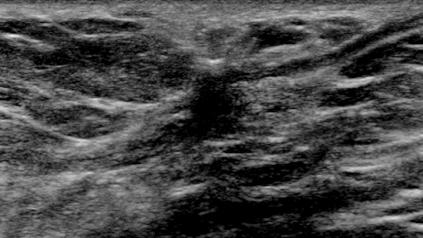}
\includegraphics[width=0.6in,height=0.6in]{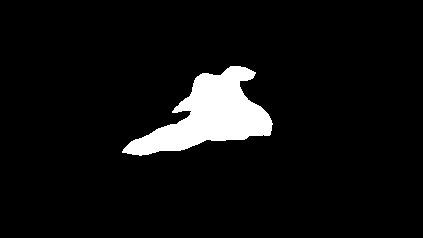}
\includegraphics[width=0.6in,height=0.6in]{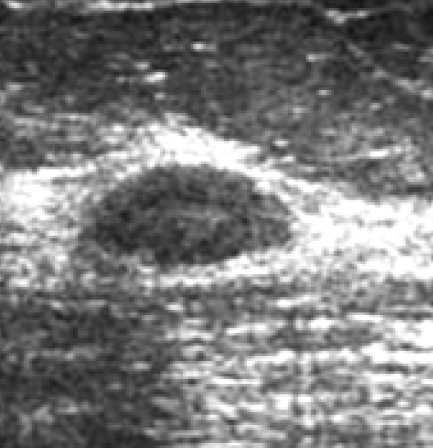}
\includegraphics[width=0.6in,height=0.6in]{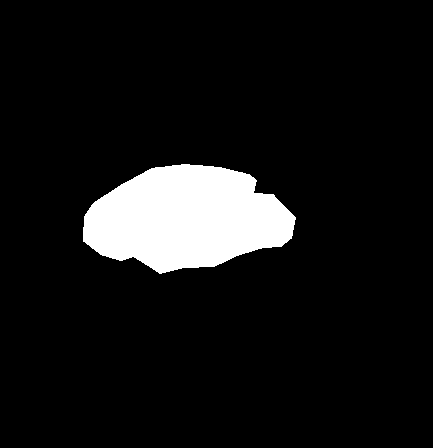}
\includegraphics[width=0.6in,height=0.6in]{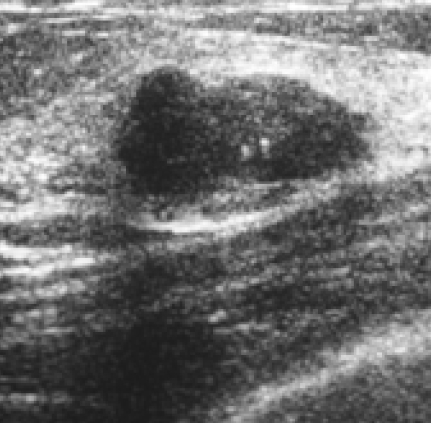}
\includegraphics[width=0.6in,height=0.6in]{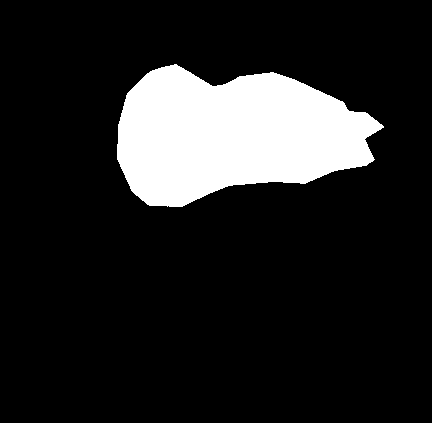} \\ \vspace{0.05in}
\includegraphics[width=0.6in,height=0.6in]{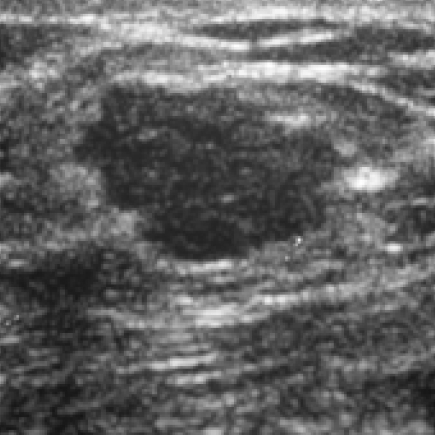}
\includegraphics[width=0.6in,height=0.6in]{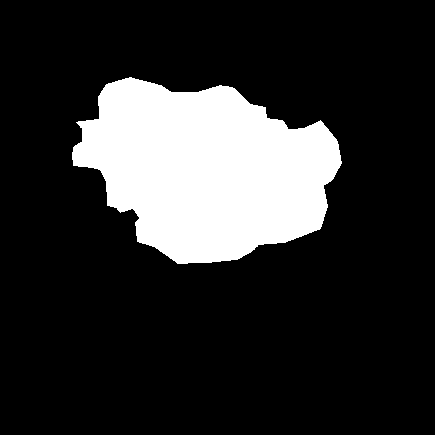}
\includegraphics[width=0.6in,height=0.6in]{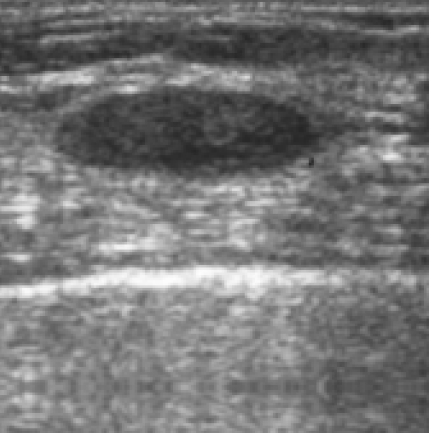}
\includegraphics[width=0.6in,height=0.6in]{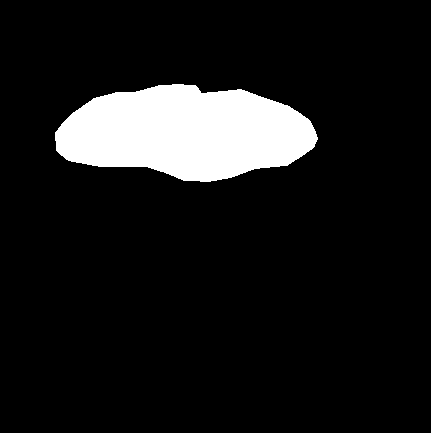}
\includegraphics[width=0.6in,height=0.6in]{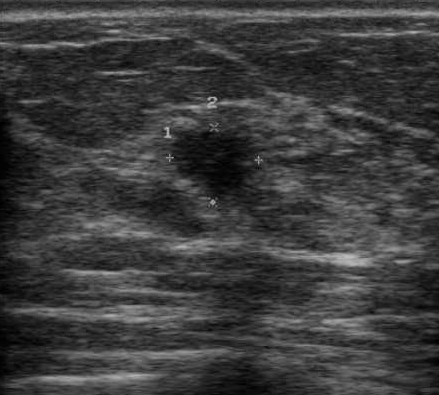}
\includegraphics[width=0.6in,height=0.6in]{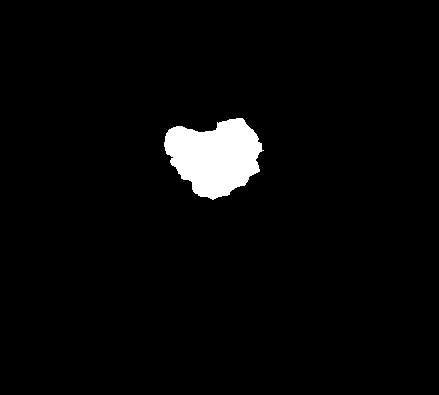}
\includegraphics[width=0.6in,height=0.6in]{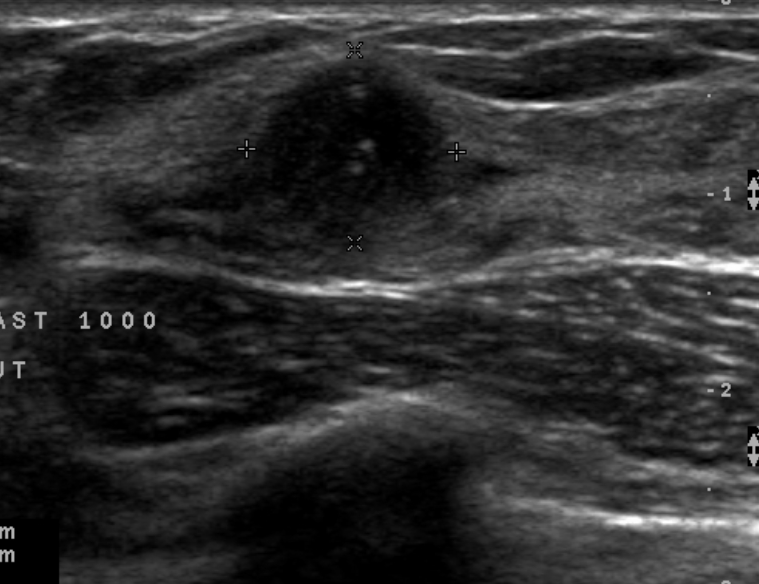}
\includegraphics[width=0.6in,height=0.6in]{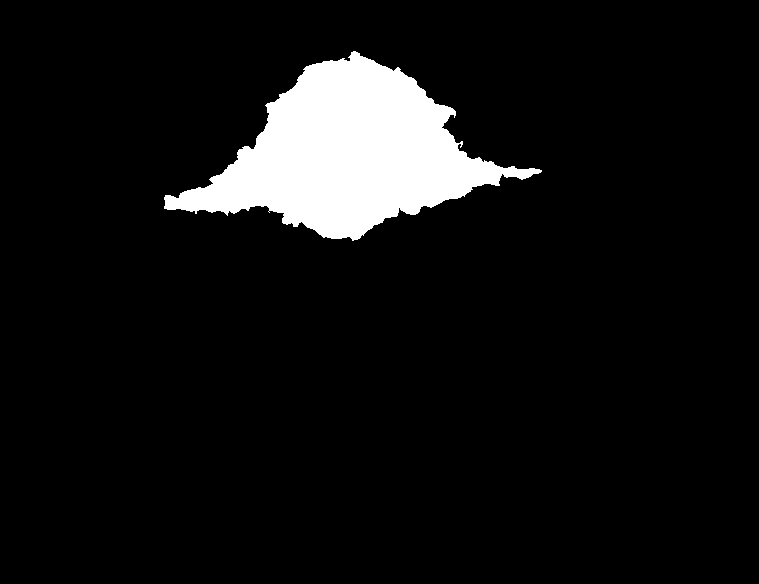}
\includegraphics[width=0.6in,height=0.6in]{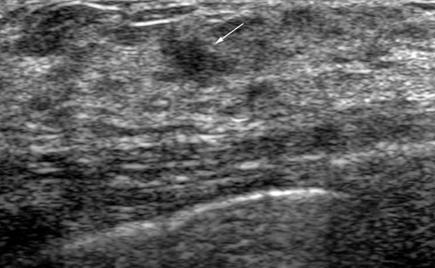}
\includegraphics[width=0.6in,height=0.6in]{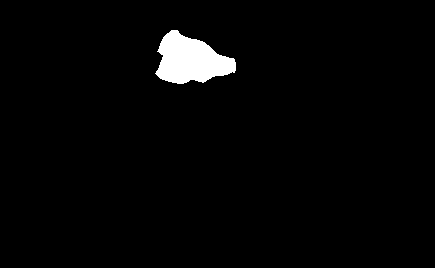} \\ \vspace{0.05in}
\includegraphics[width=0.6in,height=0.6in]{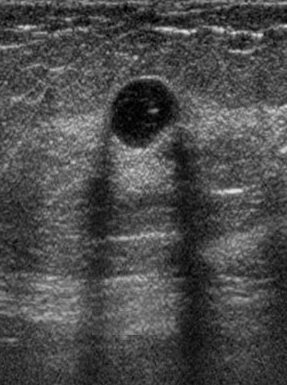}
\includegraphics[width=0.6in,height=0.6in]{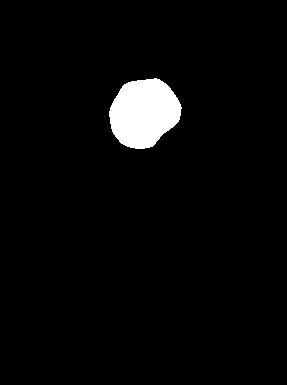}
\includegraphics[width=0.6in,height=0.6in]{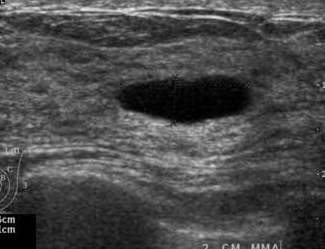}
\includegraphics[width=0.6in,height=0.6in]{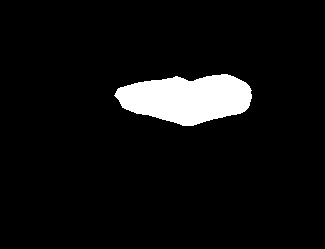}
\includegraphics[width=0.6in,height=0.6in]{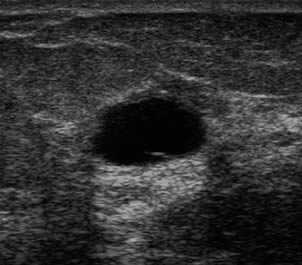}
\includegraphics[width=0.6in,height=0.6in]{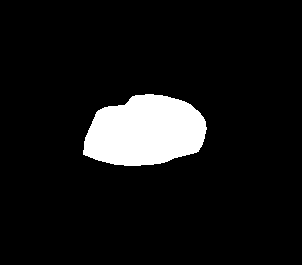}
\includegraphics[width=0.6in,height=0.6in]{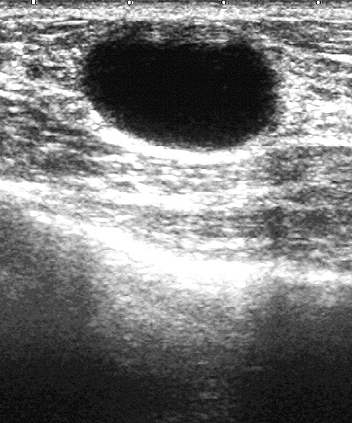}
\includegraphics[width=0.6in,height=0.6in]{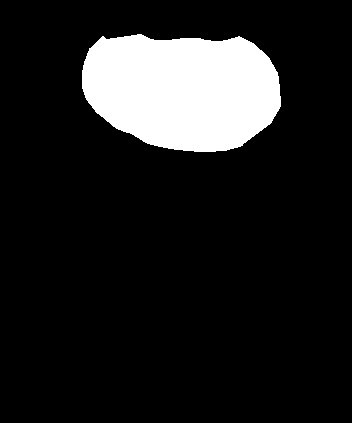}
\includegraphics[width=0.6in,height=0.6in]{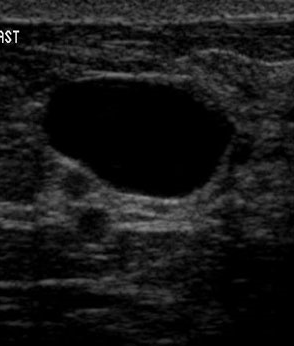}
\includegraphics[width=0.6in,height=0.6in]{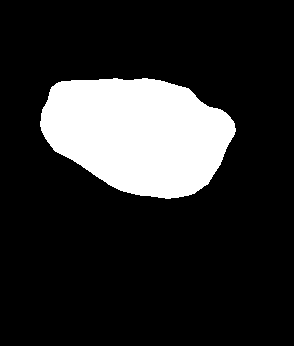} \\ \vspace{0.05in}
\includegraphics[width=0.6in,height=0.6in]{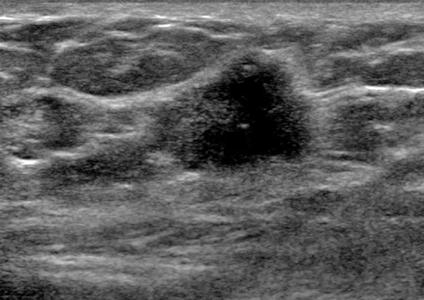}
\includegraphics[width=0.6in,height=0.6in]{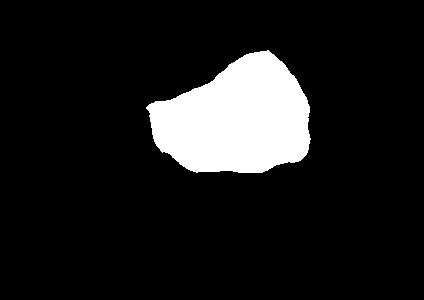}
\includegraphics[width=0.6in,height=0.6in]{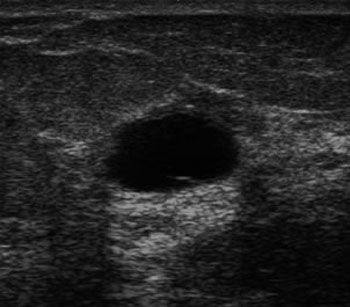}
\includegraphics[width=0.6in,height=0.6in]{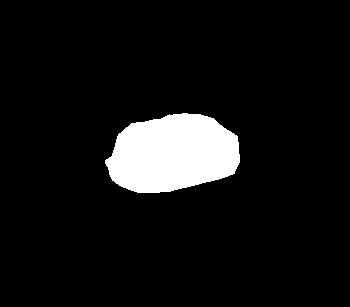}
\includegraphics[width=0.6in,height=0.6in]{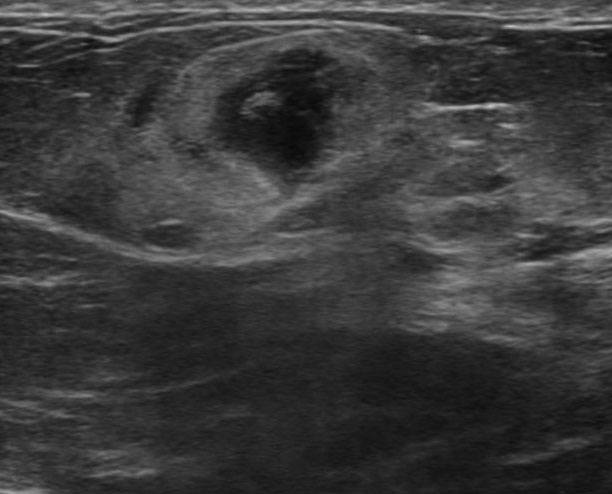}
\includegraphics[width=0.6in,height=0.6in]{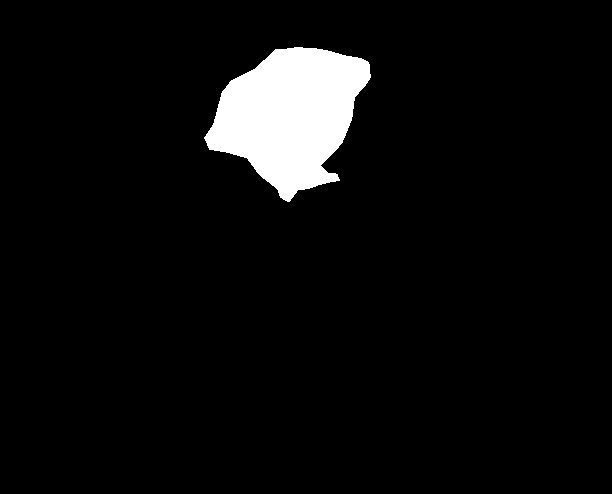}
\includegraphics[width=0.6in,height=0.6in]{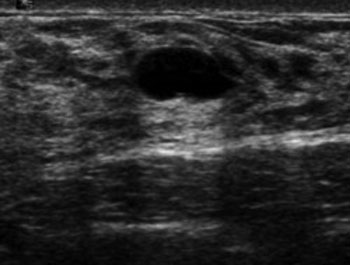}
\includegraphics[width=0.6in,height=0.6in]{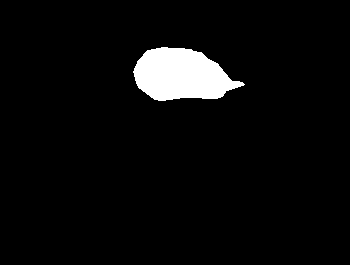}
\includegraphics[width=0.6in,height=0.6in]{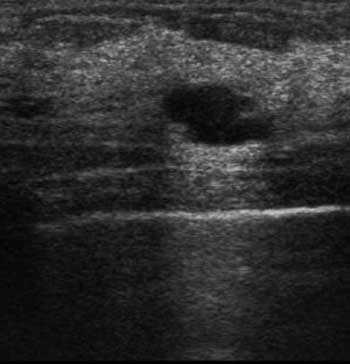}
\includegraphics[width=0.6in,height=0.6in]{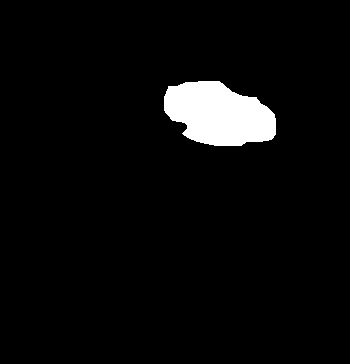} \\ \vspace{0.05in}
\includegraphics[width=0.6in,height=0.6in]{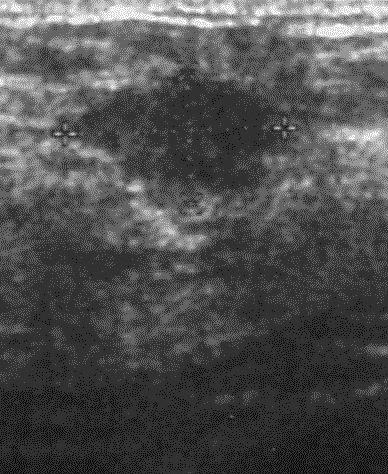}
\includegraphics[width=0.6in,height=0.6in]{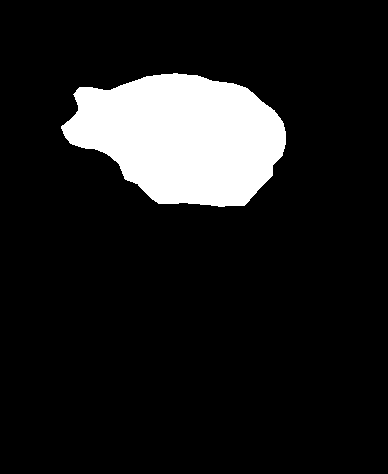}
\includegraphics[width=0.6in,height=0.6in]{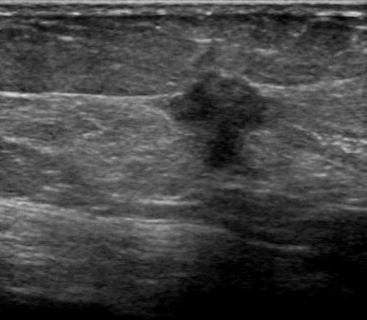}
\includegraphics[width=0.6in,height=0.6in]{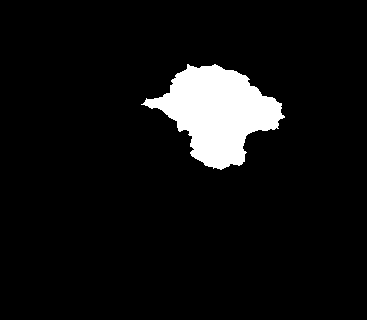}
\includegraphics[width=0.6in,height=0.6in]{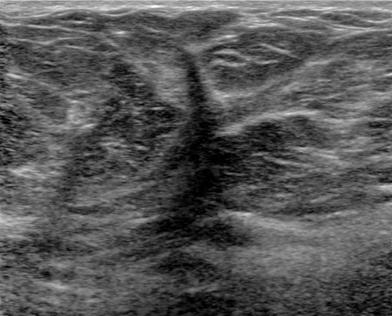}
\includegraphics[width=0.6in,height=0.6in]{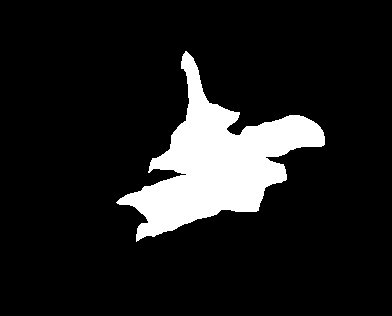}
\includegraphics[width=0.6in,height=0.6in]{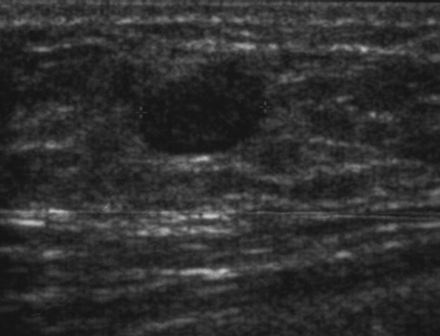}
\includegraphics[width=0.6in,height=0.6in]{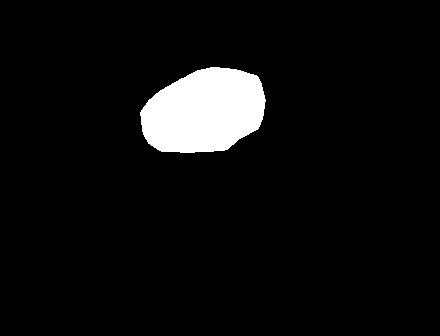}
\includegraphics[width=0.6in,height=0.6in]{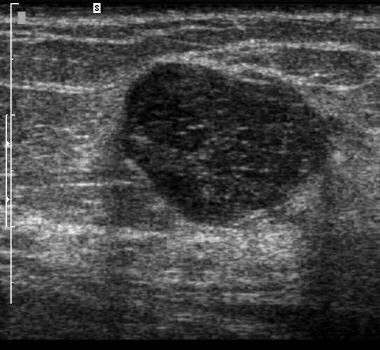}
\includegraphics[width=0.6in,height=0.6in]{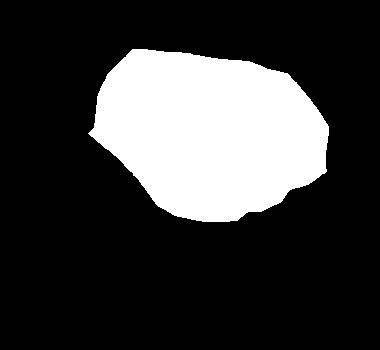} \\ \vspace{0.05in}
\includegraphics[width=0.6in,height=0.6in]{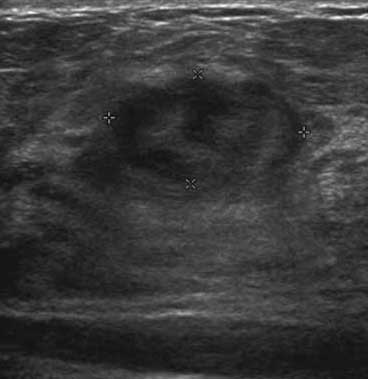}
\includegraphics[width=0.6in,height=0.6in]{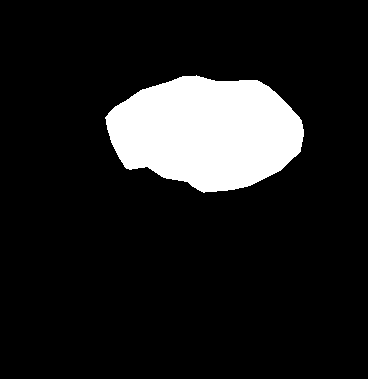}
\includegraphics[width=0.6in,height=0.6in]{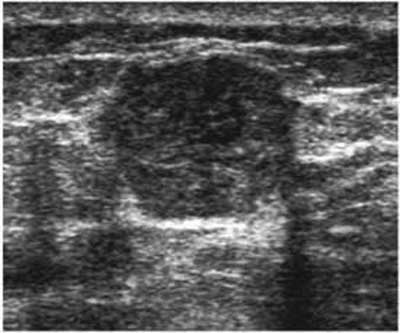}
\includegraphics[width=0.6in,height=0.6in]{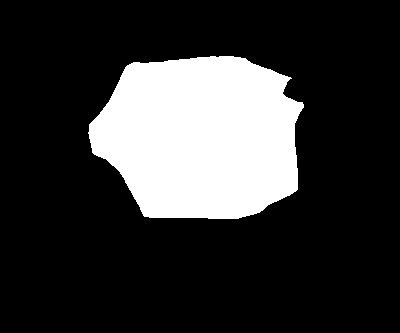}
\includegraphics[width=0.6in,height=0.6in]{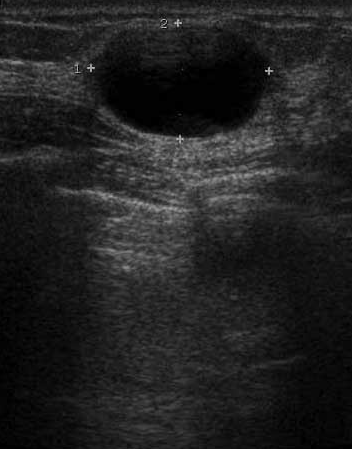}
\includegraphics[width=0.6in,height=0.6in]{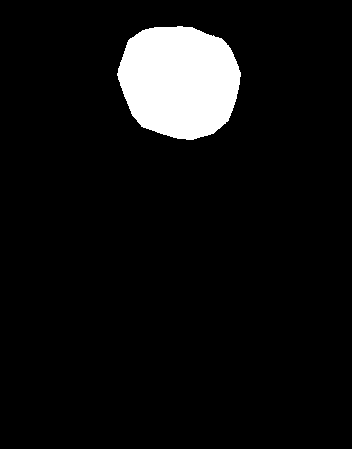}
\includegraphics[width=0.6in,height=0.6in]{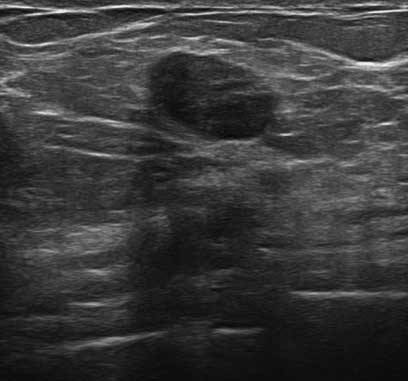}
\includegraphics[width=0.6in,height=0.6in]{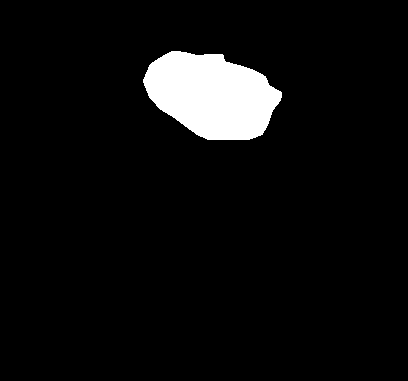}
\includegraphics[width=0.6in,height=0.6in]{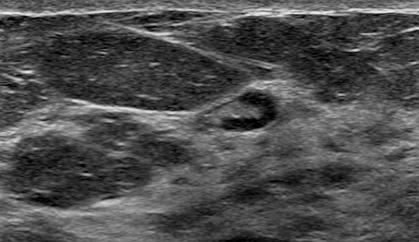}
\includegraphics[width=0.6in,height=0.6in]{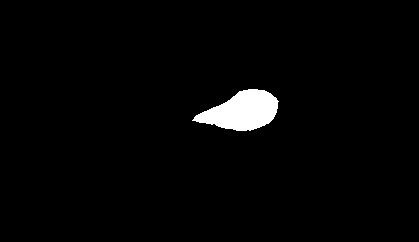}\\
\caption{Breast ultrasound scans used in our experiments. All images were segmented by an image-processing expert with extensive experience in breast lesion segmentation. Please note that some images may contain multiple ROIs. The images and their gold standard segments are available online: http://tizhoosh.uwaterloo.ca/Data/.}
\label{allImages}
\end{figure*}

The target dataset was developed from \textbf{35 breast ultrasound scans}\footnote{The images and their gold standard segments are available online: http://tizhoosh.uwaterloo.ca/Data/} that were segmented by an image-processing expert with extensive experience in breast lesion segmentation (the second author). The images, collected from the Web, are of different dimensions, ranging from $230\times 390$ to $580\times 760$ pixels (Figure \ref{allImages}, images resized for sake of illustration). These are the same images used to introduce EFIS originally \cite{EFIS}.

Ultrasound images are generally difficult to segment, primarily due to the presence of speckle noise and low level of local contrast. It should be noted that the segmentation of ultrasound actually does require a complete processing chain, (including proper preprocessing and post-processing steps). However, the purpose of using these images was solely to demonstrate that the accuracy of the segmentation can be increased with the application of SC-EFIS.

%-------------------------------------------
\subsection{Evaluation Measures}
\label{Measure}
Considering two segments $S$ (generated by an algorithm) and $G$ (the gold standard image manually created by an expert), we calculate the average of the Jaccard index $J$ (area overlap) \cite{Tan2005}:
\begin{equation}
\label{jaccardsim}
J(S,G) = \frac{|S\cap G|}{|S\cup G|},
\end{equation}
and its standard deviation $\sigma_J$. As well, the $95\%$ confidence interval (CI) of the Jaccard index $CI_J$ is calculated . Finally, we performed t-tests to validate the null hypothesis for comparing the results of a parent algorithm and its evolved version in order to establish whether any potential increase in accuracy is statistically significant. Ground-truth images $G$ were created so that the objects of interest (i.e., lesions and tumours) could be labeled as white (1) and the background as black (0). All thresholding techniques were used consistently to label object pixels in this way as this was done in EFIS.

%-----end experiment part---------
%-------the results
%-----------------SCEFIS Results------------------------
\subsection{Results}
 \label{SCEFIS}

To compare with EFIS, the SC-EFIS results are calculated for the same parent algorithms, namely for region growing (RG), global thresholding, and statistical region merging (SRM) are presented. The results are discussed with respect to rule evolution, visual inspection, accuracy verification using the Jaccard results. 

\begin{figure}[htb]
\includegraphics[width=1\columnwidth]{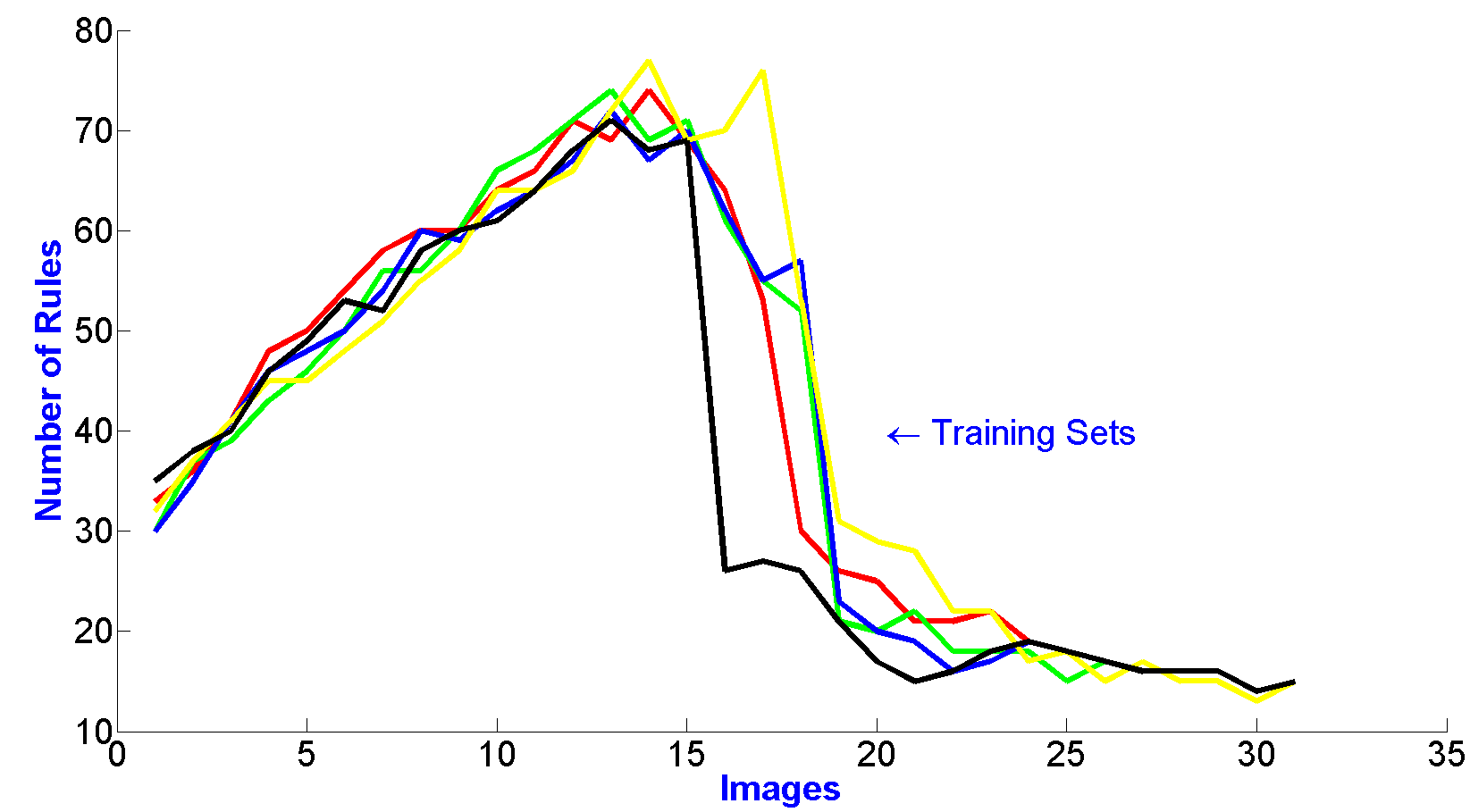}
\caption{Rule evolution for SC-EFIS for thresholding (THR): The number of rules increases first as more images are processed but then drops and seems to converge toward a lower number of rules. Each curve shows the number of rules for a separate trial/run.}
\label{Ruleevolve}
\end{figure}

 \textbf{Rule Evolution} -- Fig. \ref{Ruleevolve} indicates the change in the number of rules during the evolving of the thresholding (THR) process. The initial number of rules increases with any incoming image and then begins to decrease as additional images become available. The same behaviour was noted for SRM and RG.
 %---------------

\begin{figure}[htb]
\centering
\includegraphics[width=\columnwidth]{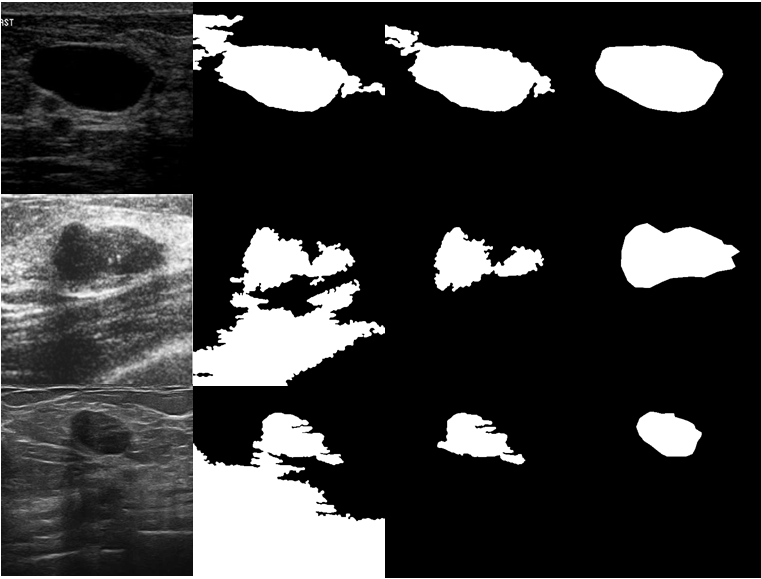}
\caption{Segmentation results: From left to right, the original image, FRG, SC-EFIS-RG, and the gold standard image.}
\label{RGSEGimages}
\end{figure}

\begin{figure}[htb]
\centering
\includegraphics[width=\columnwidth]{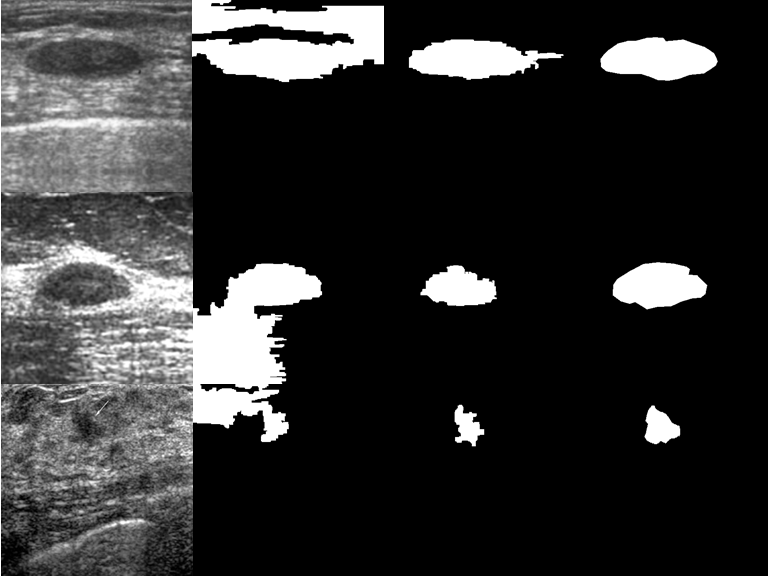}
\caption{segmentation results: From left to right, the original image, SRM, SC-EFIS-SRM, and the gold standard image.}
\label{SRMVISSEG}
\end{figure}

  \textbf{Visual Inspection} -- A visual inspection of Fig. \ref{RGSEGimages} shows that the results produced by the proposed SC-EFIS for RG represent a substantial improvement over those obtained with the FRG (fuzzy RG -- the initial fuzzy rules are used in order to estimate the similarity threshold).   A visual inspection of Fig. \ref{SRMVISSEG} reveals a significant improvement in the SC-EFIS for SRM images over the SRM ones. 

% %----------visualization Figures--------
\begin{figure}[htb]
\centering
\includegraphics[width=1\columnwidth]{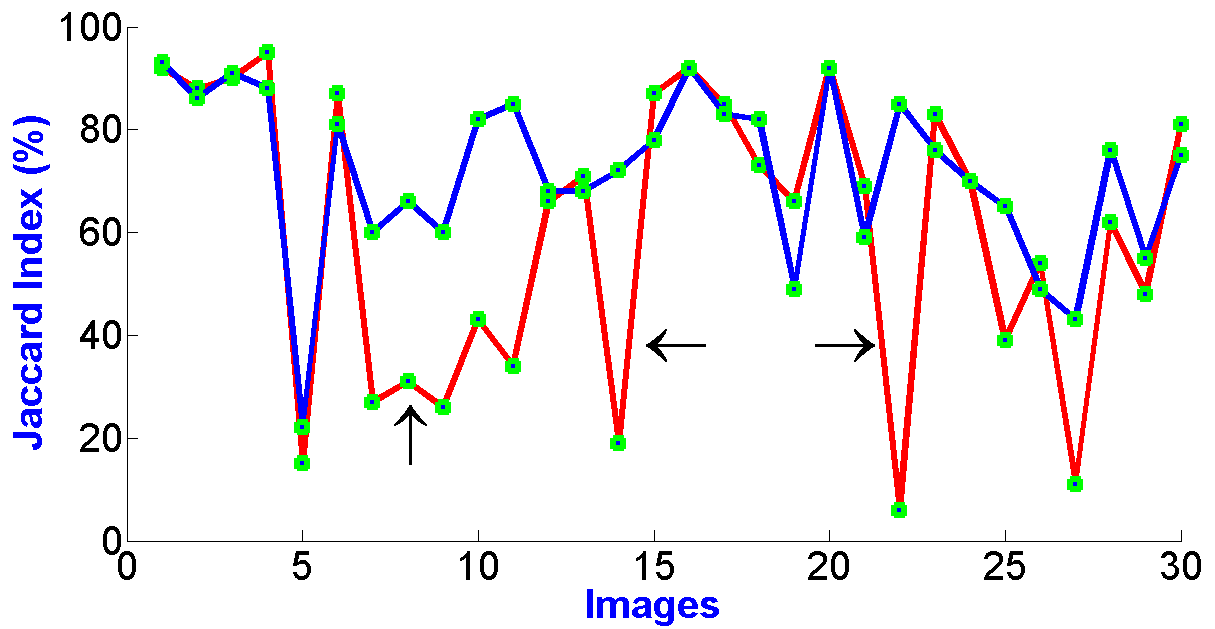}
\caption{Comparison of the Jaccard accuracy obtained with SC-EFIS-SRM (blue) and with SRM (red); arrows point to significant gaps.}
\label{JacComp}
\end{figure}

%---------------------------------
 \textbf{Accuracy Verification} -- Ten different trials/runs are presented for each method. Each run is an independent experiment involving different training and testing images. Fig. \ref{JacComp} shows the improvement in the Jaccard index of the SC-EFIS for SRM and the images for SRM with a scale = 32.

Table \ref{SCERGresults} presents a comparison of the results for the RG technique: RG results with fuzzy inference, RG results with a similarity threshold of 0.17, RG with the best similarity threshold (0.12) for the available data (RG-B), the EFIS-RG technique, and the SC-EFIS-RG. The best similarity threshold, determined only for experimental purposes, is found via exhaustive search that is impractical in real world applications. It can be seen that the results achieved with SC-EFIS are better than EFIS results in eight of ten experiments.

Table \ref{SCETHRresults} presents a comparison of the results for the global thresholding with a static (non-evolving) fuzzy system (THR) technique: the results for THR, EFIS-THR, and SC-EFIS-THR. It is clear that the SC-EFIS results surpass the EFIS ones in six of ten experiments. However, EFIS produces better results in two experiments and equivalent results in other two.

 Table \ref{SCESRMresults} presents a comparison of the results for the SRM technique: results for SRM using fuzzy inference FSRM, results for SRM with a scale = 32 (SRM), results for SRM with the best scale (64) for the available images (SRM-B) determined via exhaustive search, EFIS-SRM results, and SC-EFIS-SRM results. It can be seen that the results produced by SC-EFIS are superior to the EFIS results in five experiments, inferior in  four experiments, and equivalent for the remaining experiments. Of course, both EFIS and SC-EFIS do perform better than the parent algorithm.
 
%------------------
In general, SC-EFIS is competitive with and can even surpass EFIS with respect to the three segmentation techniques, while offering a higher level of automation.

%---------Fusion------------
\textbf{Switching/Fusion of Results} -- On the other hand, the switch/fusion technique \cite{EFIS} was re-examined for use with SC-EFIS. Table \ref{SCESWITCHresults} presents the results of switching and fusion for the same three methods, namely Niblack, SRM (scale=32), and RG (similarity = 0.17) using  EFIS (EFIS-S and EFIS-F) and using SC-EFIS (SC-EFIS-S and SC-EFIS-F). It is clear that the outcomes of EFIS and SC-EFIS are comaparable. In addition, the results with EFIS-S and SC-EFIS-S surpass those for SRM, which represents the best method. 

%%----------Tables-------------
\begin{table*}[t]
\caption{Sample results for fuzzy region growing (FRG), RG with a similarity threshold (0.17), RG-B with the best similarity threshold (0.12) (determined via exhaustive search), EFIS-RG, and SC-EFIS-RG. The null hypothesis was rejected in 10/10 runs.}
\center
\begin{tabular}{|l|l||c|c|c||c|c|}
\hline
Training &Metrics & FRG & RG & RG-B & EFIS-RG & SC-EFIS-RG \\ \hline\hline
\multirow{2}{*}{1st run}
&$J$ & 63\%& 54\% & 69\% & 68\% & 67\% \\
&$\sigma_J$& 26\% & 30\% & 21\% & 21\% & 23\%\\
&$CI_J$ & 53\%-73\%&43\%-65\%&62\%-77\%& 60\%-76\%& 58\%-75\%
\\\hline\hline
\multirow{2}{*}{2nd run}
&$J$ & 37\%& 52\% & 69\% & 63\% & 66\% \\
&$\sigma_J$& 35\% & 31\% & 19\% & 24\% & 22\%\\
&$CI_J$ & 24\%-50\%&41\%-64\%&62\%-76\%& 54\%-72\%& 57\%-74\%
\\\hline\hline
\multirow{2}{*}{3rd run}
&$J$ & 43\%& 54\% & 70\% & 65\% & 68\% \\
&$\sigma_J$& 31\% & 30\% & 21\% & 25\% & 21\%\\
&$CI_J$ & 31\%-54\%&43\%-65\%&63\%-78\%& 55\%-74\%& 61\%-76\%
\\\hline\hline
\multirow{2}{*}{4th run}
&$J$ & 33\%& 54\% & 71\% & 64\% & 66\% \\
&$\sigma_J$& 33\% & 31\% & 20\% & 23\% & 24\%\\
&$CI_J$ & 21\%-46\%&42\%-65\%&63\%-78\%& 56\%-73\%& 57\%-74\%
\\\hline\hline
\multirow{2}{*}{5th run}
&$J$ & 46\%& 54\% & 71\% & 66\% & 67\% \\
&$\sigma_J$& 32\% & 29\% & 17\% & 21\% & 20\%\\
&$CI_J$ & 34\%-58\%&43\%-65\%&64\%-77\%& 58\%-74\%& 60\%-74\%
\\\hline\hline
\multirow{2}{*}{6th run}
&$J$ & 46\%& 52\% & 69\% & 64\% & 62\% \\
&$\sigma_J$& 31\% & 30\% & 20\% & 23\% & 24\%\\
&$CI_J$ & 35\%-58\%&41\%-63\%&61\%-76\%& 55\%-73\%& 53\%-71\%
\\\hline\hline
\multirow{2}{*}{7th run}
&$J$ & 61\%& 57\% & 70\% & 67\% & 68\% \\
&$\sigma_J$& 28\% & 29\% & 21\% & 24\% & 23\%\\
&$CI_J$ & 51\%-71\%&46\%-68\%&62\%-78\%& 58\%-75\%& 59\%-76\%
\\\hline\hline
\multirow{2}{*}{8th run}
&$J$ & 56\%& 53\% & 70\% & 64\% &67\% \\
&$\sigma_J$& 30\% & 30\% & 20\% & 25\% & 23\%\\
&$CI_J$ & 45\%-67\%&42\%-64\%&62\%-78\%& 55\%-73\%& 59\%-75\%
\\\hline\hline
\multirow{2}{*}{9th run}
&$J$ & 37\%& 53\%& 70\% & 64\% & 66\% \\
&$\sigma_J$& 29\% & 31\%  & 20\% & 25\% & 23\%\\
&$CI_J$ & 26\%-48\%&41\%-64\%&63\%-78\%& 55\%-73\%& 58\%-75\%
\\\hline\hline
\multirow{2}{*}{10th run}
&$J$ & 57\%& 57\% & 71\% & 66\% & 69\% \\
&$\sigma_J$&29\% & 29\% & 18\% & 23\% & 21\%\\
&$CI_J$ & 46\%-68\%&46\%-68\%&64\%-78\%& 58\%-75\%& 61\%-77\%
\\\hline\hline
\end{tabular}
\label{SCERGresults}
\end{table*}
%%-------------
\begin{table}[htb]
\caption{Sample results for global thresholding: fuzzy thresholding (THR), EFIS-THR, and SC-EFIS-THR. The null hypothesis was rejected in 9/10 runs.}
\center
\begin{tabular}{|c|c||c||c|c|}
\hline
Training & Method & $J$ &$\sigma_J$& $CI_J$  \\ \hline\hline
\multirow{2}{*}{1st run}
 & THR 		& 58\%	&24\%	&49\%-67\%  \\
 & EFIS-THR 	& 62\%	&25\%	& 53\%-71\% \\
 & SC-EFIS-THR 	& 63\%	&23\%	& 54\%-72\% \\\hline \hline
\multirow{2}{*}{2nd run}
 & THR 		& 48\%	&33\%	&35\%-60\%  \\
  & EFIS-THR 	& 61\%	&24\%	& 52\%-70\% \\
    & SC-EFIS-THR 	& 61\%	&28\%	& 51\%-72\% \\\hline \hline
\multirow{2}{*}{3rd run}
 & THR 		& 43\%	&32\%	&31\%-55\%  \\
  & EFIS-THR 	& 63\%	&25\%	& 54\%-73\% \\
   & SC-EFIS-THR 	& 63\%	&26\%	& 53\%-72\% \\\hline \hline
\multirow{2}{*}{4th run}
 & THR 		& 23\%	&23\%	&14\%-32\%  \\
  & EFIS-THR 	& 63\%	&22\%	& 55\%-71\% \\
   & SC-EFIS-THR 	& 66\%	&21\%	& 58\%-74\% \\\hline \hline
  \multirow{2}{*}{5th run}
 & THR 		& 54\%	&26\%	&44\%-64\%  \\
  & EFIS-THR 	& 62\%	&24\%	& 53\%-71\% \\
    & SC-EFIS-THR 	& 63\%	&25\%	& 54\%-73\% \\\hline \hline
\multirow{2}{*}{6th run}
 & THR 		& 55\%	&30\%	&44\%-66\%  \\
  & EFIS-THR 	& 63\%	&23\%	& 55\%-72\% \\
    & SC-EFIS-THR 	& 64\%	&23\%	& 55\%-72\% \\\hline \hline
  \multirow{2}{*}{7th run}
 & THR 		& 38\%	&27\%	& 28\%-48\%  \\
  & EFIS-THR 	& 60\%	&24\%	& 51\%-69\% \\
    & SC-EFIS-THR 	& 59\%	&26\%	& 49\%-69\% \\\hline \hline
\multirow{2}{*}{8th run}
 & THR 		& 52\%	&24\%	&43\%-62\%  \\
  & EFIS-THR 	& 62\%	&21\%	& 54\%-70\% \\
   & SC-EFIS-THR 	& 63\%	&21\%	& 55\%-70\% \\\hline \hline
\multirow{2}{*}{9th run}
 & THR 		& 39\%	&31\%	&28\%-51\%  \\
  & EFIS-THR 	& 63\%	&23\%	& 54\%-73\% \\
   & SC-EFIS-THR 	& 65\%	&21\%	& 57\%-73\% \\\hline \hline
\multirow{2}{*}{10th run}
 & THR 		& 44\%	&25\%	&34\%-53\%  \\
  & EFIS-THR 	& 58\%	&26\%	& 48\%-68\% \\
   & SC-EFIS-THR 	& 57\%	&26\%	& 47\%-67\% \\\hline
     \hline
\end{tabular}

\label{SCETHRresults}
\end{table}
\begin{table*}[bth]
\caption{Sample results for fuzzy statistical region merging (FSRM), SRM with the default scale (32), SRM-B with the best scale (64) (determined via exhaustive search), EFIS-SRM, and SC-EFIS-SRM. The null hypothesis was rejected in 10/10 runs.}
\center
\begin{tabular}{|l|l||c|c|c||c|c|}
\hline
Training &Metrics & FSRM & SRM & SRM-B & EFIS-SRM & SC-EFIS-SRM \\ \hline\hline
\multirow{2}{*}{1st run}
&$J$ & 64\%& 60\% & 72\% & 71\% & 72\% \\
&$\sigma_J$& 24\% & 28\% & 21\% & 19\% & 17\%\\
&$CI_J$ & 55\%-73\%&50\%-71\%&64\%-79\%& 64\%-78\%& 65\%-78\%
\\\hline \hline
\multirow{2}{*}{2nd run}
&$J$ & 66\%& 60\% & 68\% & 69\% & 67\% \\
&$\sigma_J$& 25\% & 27\% & 24\% & 22\% & 20\%\\
&$CI_J$ & 57\%-76\%&50\%-70\%&59\%-76\%& 61\%-77\%& 60\%-75\%
\\\hline\hline
\multirow{2}{*}{3rd run}
&$J$ & 63\%& 61\% & 70\% & 67\% & 69\% \\
&$\sigma_J$& 25\% & 28\% & 22\% & 24\% & 18\%\\
&$CI_J$ & 53\%-72\%&50\%-71\%&62\%-78\%& 58\%-76\%& 62\%-76\%
\\\hline\hline
\multirow{2}{*}{4th run}
&$J$ & 57\%& 59\% & 69\% & 71\% & 71\% \\
&$\sigma_J$& 29\% & 30\% & 24\% & 21\% & 19\%\\
&$CI_J$ & 46\%-67\%&48\%-70\%&60\%-78\%& 63\%-79\%& 64\%-78\%
\\\hline\hline
\multirow{2}{*}{5th run}
&$J$ & 42\%& 59\% & 68\% & 67\% & 68\% \\
&$\sigma_J$& 33\% & 29\% & 24\% & 23\% & 22\%\\
&$CI_J$ & 30\%-54\%&49\%-70\%&59\%-77\%& 59\%-76\%& 60\%-77\%
\\\hline\hline
\multirow{2}{*}{6th run}
&$J$ & 63\%& 60\% & 69\% & 69\% & 68\% \\
&$\sigma_J$& 26\% & 28\% & 22\% & 21\% & 20\%\\
&$CI_J$ & 53\%-73\%&49\%-70\%&61\%-77\%& 61\%-76\%& 61\%-76\%
\\\hline\hline
\multirow{2}{*}{7th run}
&$J$ & 55\%& 61\% & 70\% & 70\% & 70\% \\
&$\sigma_J$& 30\% & 29\% & 23\% & 22\% & 20\%\\
&$CI_J$ & 44\%-67\%&50\%-72\%&62\%-79\%& 62\%-79\%& 63\%-78\%
\\\hline\hline
\multirow{2}{*}{8th run}
&$J$ & 67\%& 59\% & 70\% & 68\% &69\% \\
&$\sigma_J$& 19\% & 28\% & 22\% & 22\% & 20\%\\
&$CI_J$ & 60\%-74\%&48\%-69\%&62\%-78\%&60\%-76\%& 62\%-76\%
\\\hline\hline
\multirow{2}{*}{9th run}
&$J$ & 47\%& 59\% & 69\% & 71\% & 67\% \\
&$\sigma_J$& 31\% & 30\% & 24\% & 22\% & 24\%\\
&$CI_J$ & 36\%-59\%&47\%-70\%&60\%-78\%& 63\%-79\%& 58\%-76\%
\\\hline\hline
\multirow{2}{*}{10th run}
&$J$ & 64\%& 61\% & 69\% & 68\% & 71\% \\
&$\sigma_J$& 28\% & 29\% & 24\% & 23\% & 19\%\\
&$CI_J$ & 54\%-74\%&51\%-72\%&60\%-78\%& 60\%-77\%& 64\%-78\%
\\\hline\hline
\end{tabular}

\label{SCESRMresults}
\end{table*}
%
%%-----------------
\begin{table*}[tb]
\caption{Accuracy of switching and fusion for three methods: Niblack, SRM, and RG using EFIS and SC-EFIS: Each dataset had 30 images for training and 5 images for testing.}
\centering
\begin{tabular}{|c||c|c|c||c|c|c|c|c|c|}
\hline
Dataset &	Niblack  &	SRM &	RG &	EFIS-S & EFIS-F &SC-EFIS-S&SC-EFIS-F \\ \hline
1	& 76\%	& 68\%	 & 50\% 	& 77\% & 77\%&76\%&65\%\\
2	& 52\%	& 55\%	& 48\%	& 53\% & 53\%&62\%&52\% \\
3	& 77\%	& 74\%	& 72\%	& 80\% & 72\%&80\%&81\%\\
4	& 74\%	& 57\%	& 55\%	& 55\% & 56\%&65\%&66\%\\
5	& 43\%	& 33\%	& 33\%	& 36\% & 36\%&34\%&28\%\\
6	& 59\%	& 59\%	& 62\%	& 62\% & 61\% &61\%&57\%\\
7	& 55\%	& 82\%	& 80\%	& 81\% & 78\%&62\%&78\%\\
8	& 62\%	& 62\%	& 58\%	& 66\% & 65\%&63\%&58\%\\
9	& 68\%	& 64\%	& 63\%	& 76\% & 70\%&73\%&69\%\\
10	& 59\%	& 90\%	& 89\%	& 79\% & 79\%&76\%&90\%\\ \hline\hline
$m$ & 62.3\%     & 64.5\%     &  61.0\%     & \cellcolor[gray]{0.8}  66.5\%  & \cellcolor[gray]{0.8}  64.6\%  & \cellcolor[gray]{0.8}  64.9\%  & \cellcolor[gray]{0.8}  64.3\%\\
$\sigma$ & 11\%     & 16\%     &  16\%     & \cellcolor[gray]{0.8}  15\%  & \cellcolor[gray]{0.8}  13\% &  \cellcolor[gray]{0.8}  14\%  & \cellcolor[gray]{0.8}  17\%\\
\hline
\end{tabular}

\label{SCESWITCHresults}
\end{table*}
Table \ref{SCCOMPresults} enables a comparison of EFIS and SC-EFIS results for global thresholding with different global and local thresholding techniques. The data listed are taken form three experiments selected from Table \ref{SCETHRresults}. It is clear that, in the three experiments, EFIS and SC-EFIS provide outcomes that are more accurate than those produced with the non-evolutionary thresholding techniques.

\begin{table*}[htb]
\center
\caption{Comparison of EFIS, SC-EFIS, and 4 other global thresholding technique as well as one local thresholding method (\cite{tizhoosh2005image,Tizhoosh2008inbook, nib, Kittler1986,Huang1995}): Average and standard deviation of the Jaccard index $J \pm \sigma_J$ and 95\% confidence interval $CI_J$. The MAA indicates the maximum achievable accuracy determined via exhaustive search and through comparison with gold standard images; no global thresholding method can achieve higher accuracies than MAA.}
\begin{tabular}{|c|l|c|c|c|}
\hline
Run	& Method	& $J \pm \sigma_J$ & $CI_J$  \\ \hline
	& \cellcolor[gray]{0.8} MAA 		&  \cellcolor[gray]{.8} 79\%$\pm$12\% & \cellcolor[gray]{.8} [75\%	84\%]  \\
	& \textbf{EFIS-THR}	&  \textbf{62\%$\pm$25\%} & \textbf{[53\%	71\%]} \\
	& \textbf{SC-EFIS-THR}	&  \textbf{63\%$\pm$23\%} & \textbf{[54\%	72\%]} \\
	& Niblack	(local)		&  56\%$\pm$24\% & [47\%	65\%] \\
1	& Huang			&  45\%$\pm$27\% & [35\%	55\%] \\
	& Kittler			&  39\%$\pm$32\% & [27\%	51\%] \\
	& Tizhoosh		&  35\%$\pm$32\% & [23\%	47\%] \\
	& Otsu     			&  28\%$\pm$25\% & [18\%	37\%] \\ \hline
	& \cellcolor[gray]{0.8}  MAA 		&  \cellcolor[gray]{0.8}  79\%$\pm$11\% & \cellcolor[gray]{0.8}  [75\%	83\%]  \\
	& \textbf{EFIS-THR}	&  \textbf{60\%$\pm$24\%} & \textbf{[51\%	69\%]} \\
		& \textbf{SC-EFIS-THR}	&  \textbf{59\%$\pm$26\%} & \textbf{[49\%	69\%]} \\
	& Niblack	(local)		&  57\%$\pm$25\% & [48\%	66\%] \\
2	& Huang			&  44\%$\pm$29\% & [34\%	55\%] \\
	& Kittler			&  41\%$\pm$31\% & [29\%	52\%] \\
	& Tizhoosh		&  38\%$\pm$32\% & [26\%	50\%] \\
	& Otsu     			&  29\%$\pm$25\% & [19\%	38\%] \\ \hline
	& \cellcolor[gray]{0.8}  MAA 		&  \cellcolor[gray]{0.8}  79\%$\pm$12\% & \cellcolor[gray]{0.8}  [74\%	83\%]  \\
	& \textbf{EFIS-THR}	&  \textbf{63\%$\pm$23\%} & \textbf{[54\%	71\%]} \\
			& \textbf{SC-EFIS-THR}	& \textbf{65\%$\pm$21\%} & \textbf{[57\%	73\%]} \\
	& Niblack	(local)		&  59\%$\pm$24\% & [49\%	68\%] \\
3	& Huang			& 46\%$\pm$27\% & [35\%	56\%] \\
	& Kittler			&  41\%$\pm$33\% & [29\%	53\%] \\
	& Tizhoosh		&  35\%$\pm$33\% & [23\%	48\%] \\
	& Otsu     			&  28\%$\pm$23\% & [20\%	37\%] \\
\hline
\end{tabular}
\label{SCCOMPresults}
\end{table*}

%*************************************************
%*************************************************
%*************************************************
\section{Conclusions}
\label{CON}
Most image segmentation techniques involve multiple parameters that must be tuned in order to achieve maximum segmentation accuracy. Evolving fuzzy image segmentation (EFIS) has been recently proposed to provide evolving and user-oriented adjustment for medical image segmentation. EFIS is a generic segmentation scheme that relies on user feedback in order to improve the quality of segmentation. Its evolving nature makes this approach attractive for applications that incorporate high-quality user feedback, such as in medical image analysis. 
 However, EFIS entails some limitations, such as parameters that must be selected prior to the running of the algorithm and the lack of an automated feature selection component. These drawbacks restrict the use of EFIS to specific categories of images. 
 An improved version of EFIS, called self-configuring EFIS (SC-EFIS) was proposed in this paper. SC-EFIS is a generic image segmentation scheme that does not require setting of some parameters, such as number of features or detecting a region of interest. SC-EFIS operates with the data available and extracts major parameters necessary for its operation from those data. A comparison of the SC-EFIS results with those obtained with EFIS demonstrates the comparable accuracy of both schemes with SC-EFIS offering a much higher level of automation.

%--------bib and end document---------
\bibliographystyle{siam}
\bibliography{ref4}
\end{document}